\documentclass[runningheads]{llncs}

\usepackage{eccv}

\usepackage{eccvabbrv}

\usepackage{graphicx}
\usepackage{booktabs}
\usepackage{multirow}
\usepackage{xcolor}
\usepackage{makecell}

\usepackage{algorithm}
\usepackage{algpseudocode} 
\usepackage{capt-of}       

\usepackage[accsupp]{axessibility}  

\usepackage{hyperref}

\usepackage{orcidlink}

\newcommand{\PPS}{\textsc{PPS}}
\newcommand{\PSP}{\textsc{PSP}}
\newcommand{\BoN}{\textsc{BoN}}
\newcommand{\IR}{\textsc{IR}}
\newcommand{\HPS}{\textsc{HPS}}

\begin{document}

\title{Inference-Time Scaling of Diffusion Models via Progressive Seed Pruning}

\author{Rog\'erio Guimar\~aes\inst{1}\orcidlink{0000-0002-1693-4241} \and
Pietro Perona\inst{1}\orcidlink{0000-0002-7583-5809} }

\authorrunning{R.~Guimar\~aes and P.~Perona}

\institute{California Institute of Technology, Pasadena CA 91125, USA
\email{\{rogerio,perona\}@caltech.edu}\\
\url{https://vision.caltech.edu}}

\maketitle

\begin{abstract}
Diffusion and flow-matching models dominate conditional image generation, yet inference-time scaling for these models is far less developed than for autoregressive language models. Because final quality is highly sensitive to the initial noise seed, many approaches spend extra compute on seed search or resampling under a black-box reward, but typically maintaining a constant memory footprint throughout inference. We show that relaxing this constraint enables an underexplored inference-time scaling axis: by front-loading exploration, evaluating many seeds early, and pruning aggressively, we can use a fixed compute budget more effectively. \emph{Progressive Seed Pruning} (\PSP) scores intermediate denoised estimates and progressively narrows the candidate set so that only promising trajectories are fully denoised, while keeping the total number of model evaluations fixed. Across diffusion and flow-matching backbones, \PSP \ consistently improves reward-guided selection and achieves higher GenEval scores (automated) and better human evaluation on prompt-alignment than best-of-$N$, importance-sampling, and tree-search baselines at matched compute. Project page: \href{https://www.vision.caltech.edu/psp/}{\texttt{vision.caltech.edu/psp}}. Code: \href{https://github.com/rogerioagjr/psp}{\texttt{github.com/rogerioagjr/psp}}

\keywords{Diffusion models \and flow matching \and inference-time scaling \and reward-guided generation \and prompt alignment \and particle filtering}
\end{abstract}

\section{Introduction}
\label{sec:intro}

\begin{figure*}[t]
  \centering
  \includegraphics[width=\textwidth]{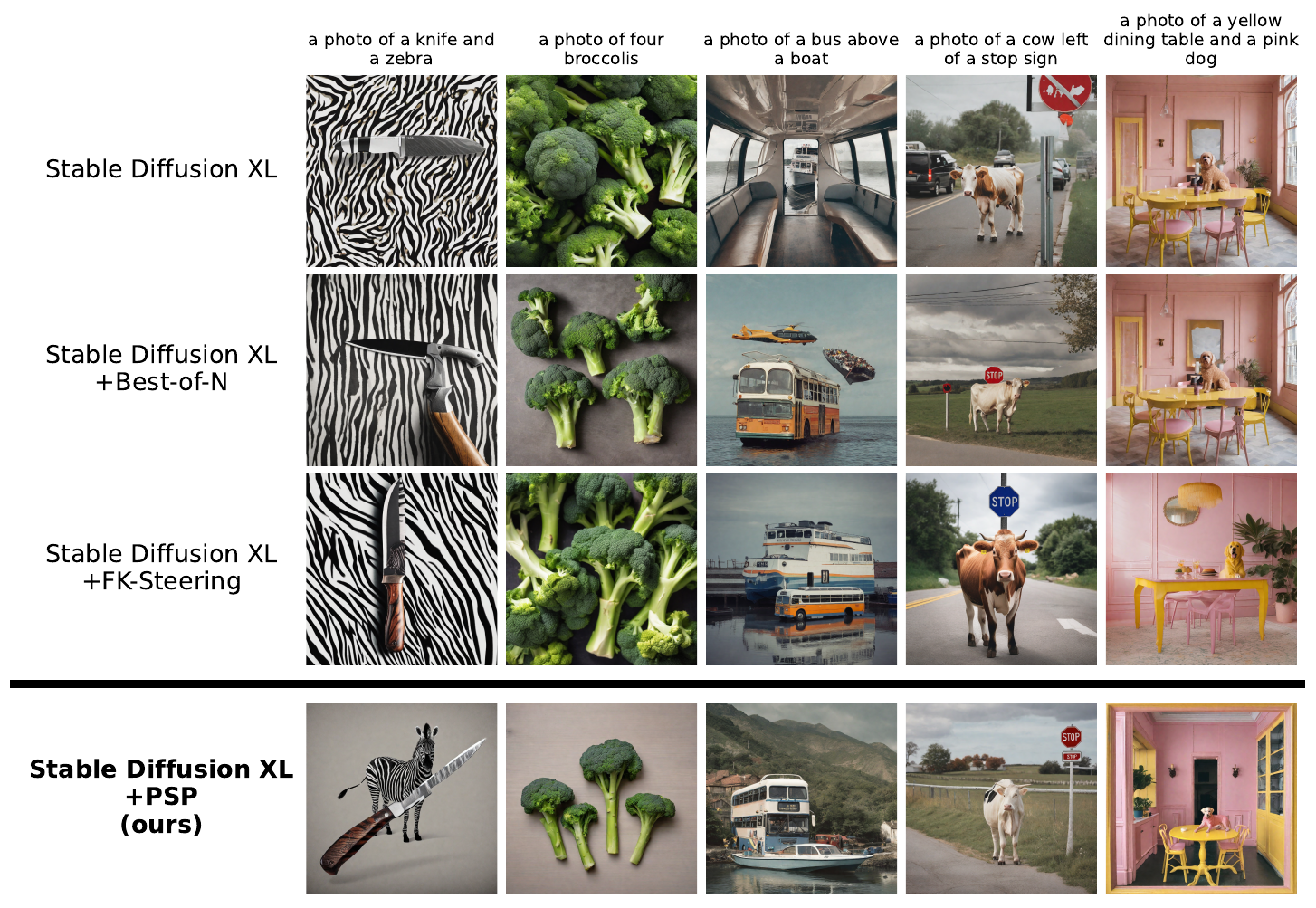}
  \caption{\textbf{Examples of image generation improvement with \PSP.}
  SDXL generations using a standard sampler and three gradient-free inference-time scaling strategies under a fixed compute multiplier $\bar{N}=4$: \BoN, FK-Steering (importance-sampling based), and our \PSP. We show examples from GenEval prompts in which \PSP \ improves on \BoN \ and FK-Steering, satisfying prompt constraints (full results in \cref{tab:main_models_samplers}).}
  \label{fig:teaser_methods_grid}
\end{figure*}

Diffusion \cite{sohl-dickstein_deep_2015, ho_denoising_2020, song_scorebased_2021} and Flow Matching models \cite{liu_flow_2022, lipman_flow_2023} have become the default engines for conditional image and video generation. At the same time, a central lesson from large language models is that \emph{inference-time scaling} can be as important as training-time scaling: additional compute at test time, via search, sampling, and verification, can yield large quality gains \cite{yao_tree_2023,besta_graph_2024,cobbe_training_2021,yao_webshop_2023,valmeekam_planning_2023}. For diffusion-style generators, the dominant knob has been to increase denoising steps, but this is often an inefficient way to spend extra compute.

A growing body of work points to a different lever: for modern text-to-image systems, the \emph{random seed} (initial noise) can strongly influence generation quality \cite{qi_not_2024, xu_good_2025}. This motivates inference-time strategies that treat generation as \emph{search} over seeds guided by a black-box reward model. Early work showed that Best-of-N (\BoN) already scales better than using more timesteps \cite{ma_inferencetime_2025}, but it fully denoises many candidates that are unlikely to win.

Recent methods use \emph{intermediate} reward signals to allocate compute more efficiently, via importance sampling \cite{singhal_general_2025} or tree search \cite{li_dynamic_2025, zhang_inferencetime_2025, oshima_inferencetime_2025}. These approaches are gradient-free and require no backbone finetuning, but typically enforce a \emph{fixed} number of parallel samples (constant memory) throughout the trajectory. In contrast, particle filtering has long studied variable particle counts and early pruning \cite{fox_kldsampling_2001}, but this idea has not been thoroughly stress-tested for modern diffusion-style image generation.

In this paper we revisit a simple question: \emph{if we relax the constant-memory constraint, how should we allocate compute over the denoising process?} Intermediate reward estimates often become informative relatively early. That suggests a straightforward strategy: begin with a large pool of seeds, advance them only until reward signals become meaningful, prune them progressively, and reserve the remaining denoising budget for a shrinking set of promising candidates.

We study a simple instantiation of this idea, \emph{Progressive Seed Pruning} (\PSP). \PSP\ begins from a large pool of noise seeds, scores intermediate denoised estimates, and prunes on a predetermined schedule so that only promising trajectories receive the remaining denoising budget. This front-loads exploration and improves reward-guided selection over prior inference-time scaling baselines while keeping the total number of model evaluations fixed. Using a fixed schedule also aligns with deployment: predetermined prune points and survivor counts make each interval’s memory footprint and runtime predictable, facilitating allocation on its ideal setting: multi-GPU servers with elastic. Taken together, our results suggest that \emph{variable particle populations} are a simple, general, and deployment-friendly design principle for inference-time scaling in modern generative models.

\paragraph{Contributions.}
\begin{enumerate}
\item We identify \textbf{constant-memory inference} as an unnecessary restriction and show that \textbf{time-varying particle counts} are a strong inference-time scaling principle for diffusion/flow models under black-box rewards when used to front-load compute and consider a larger pool of initial noise seeds.
\item We demonstrate consistent gains of \PSP \ over \BoN, importance-sampling, and tree-search baselines at matched compute across diffusion and flow-matching backbones on {\bf both automated metrics and human evaluation}, and show that \textbf{performance scales with compute}.
\item We show that determinism enables \textbf{offline schedule search}: pruning schedules can be simulated from cached intermediate rewards without rerunning the generator, enabling rapid per-task adaptation.
\end{enumerate}

\section{Related Work}
\label{sec:related}

\subsection{Using Rewards to Steer Diffusion}
A common way to use rewards for conditional generation is \emph{gradient-based guidance} \cite{chung_diffusion_2024, bansal_universal_2023}, which backpropagates gradients of a differentiable objective to bias the sampling trajectory or the initial noise \cite{tang_inferencetime_2024} towards higher reward. While effective, this requires differentiable rewards and adds significant inference overhead due to gradient computation and extra evaluations, making it less practical when rewards are black-box or expensive.

Another option is to directly \emph{finetune} the generative backbone to maximize a reward with Reinforcement Learning \cite{black_training_2024, yang_using_2024} or preference objective \cite{wallace_diffusion_2023}. This can improve default sampling without requiring inference-time search, but it comes at the cost of additional training of large generative models and the associated engineering overhead. In contrast, \PSP\ is training-free and gradient-free, making it straightforward to apply off-the-shelf across rewards and across both diffusion and flow-matching backbones.

\subsection{Scaling Inference Compute with Gradient-Free Rewards}
Recent methods use \emph{intermediate} signal from black-box reward functions to scale compute at inference time efficiently by running multiple samples in parallel allocating compute to the most promising ones. FK-Steering \cite{singhal_general_2025} views diffusion inference through the lens of a \emph{Feynman--Kac interacting particle system} \cite{delmoral_feynmankac_2004, carmona_interacting_2009} and generalizes importance-sampling approaches, like TDS \cite{wu_practical_2024} and SVDD \cite{li_derivativefree_2024}. It maintains a population of particles and periodically reweights and resamples them using potentials derived from intermediate reward estimates.

Another prominent line of work casts inference-time sample selection as a tree search over the denoising trajectory \cite{zhang_inferencetime_2025, li_dynamic_2025, ramesh_testtime_2025}. We choose DSearch \cite{li_dynamic_2025} as their main representative in our \emph{fixed-budget} per generation setting. It performs a beam-style search over partial trajectories, shrinking the beam over time while increasing branching (multiple children per candidate) to better estimate value from intermediate states. 

Other related approaches pursue a different objective by allocating \emph{adaptive per-prompt compute}, revisiting earlier steps for refinement and extending inference until a stopping rule indicates the generation is satisfactory \cite{zhang_inferencetime_2025, lee_adaptive_2025}.

A shared assumption in both FK-Steering and DSearch is a \emph{constant memory budget} during inference: the maximum number of simultaneously explored samples per step is kept roughly fixed. Our work departs from this design choice by relaxing the constant-memory constraint and instead using early pruning to spend more compute on considering a larger set of initial seeds.

\subsection{Adaptive Particle Counts in Particle Filtering and Pruning}
In classical particle filtering and sequential Monte Carlo, it is common to propagate a fixed number of particles at every time step, even though the complexity of the target distribution can vary significantly over time. A long-standing line of work therefore studies \emph{adaptive} particle counts to allocate computation where it is most needed. KLD-sampling adapts the number of particles to satisfy a prescribed approximation-quality bound measured via a Kullback--Leibler criterion, using more particles in ambiguous phases and fewer once the posterior concentrates \cite{fox_kldsampling_2001}. More recent ``adaptive $N$'' particle filters adjust particle counts based on online predictive statistics and provide guarantees showing how error bounds track these updates over time~\cite{elvira_performance_2021}. In parallel, the Multi-Armed Bandit literature heavily relies on \emph{Successive Halving} \cite{karnin_almost_2013, jamieson_nonstochastic_2015}, an algorithm that optimally distributes a fixed compute budget by evaluating a massive pool of configurations and iteratively discarding the worst-performing half. This is similar to our default pruning schedule (\cref{sec:tuning}), but \PSP \ is more flexible and allows for searching for optimal schedules for a given task, rewards, and prompts.

Our work can be seen as a simple application of these adaptive-particle and early-pruning ideas to reward-guided inference for diffusion image generation. Although the particle-filtering literature explores richer adaptive strategies, we show that a fixed, predetermined pruning schedule already captures much of the benefit of adaptive particle counts for conditional image generation.

The allocation principle itself is classical, yet prior reward-guided diffusion methods largely \emph{fix parallelism}. The closest exception is the adaptive sampler of \cite{kim_inferencetime_2025}, termed \emph{rollover budget forcing}, which spends compute unevenly over the course of inference. However, they allocate in the opposite direction to us: instead of considering a larger pool of seeds early, they conserve budget early and resample more heavily later. For tasks like prompt alignment, this misplaces compute where it matters least, since the important features are coarse and thus largely fixed early in the generative process \cite{park_understanding_2023}, a property already exploited when using diffusion backbones for perception tasks \cite{luo_diffusion_2023, guimaraes_diffusion_2026}. Therefore, their adaptive sampler alone gains little over other sampling and search methods. By contrast, we show that relaxing constant memory to expand the pool of initial seeds early and prune over time is a surprisingly strong baseline for modern diffusion \emph{and} flow-matching generators.

\section{Progressive Seed Pruning}
\label{sec:method}

\begin{figure}[t]
  \centering
  \newcommand{\trajwidth}{0.95\columnwidth}
  \includegraphics[width=\trajwidth]{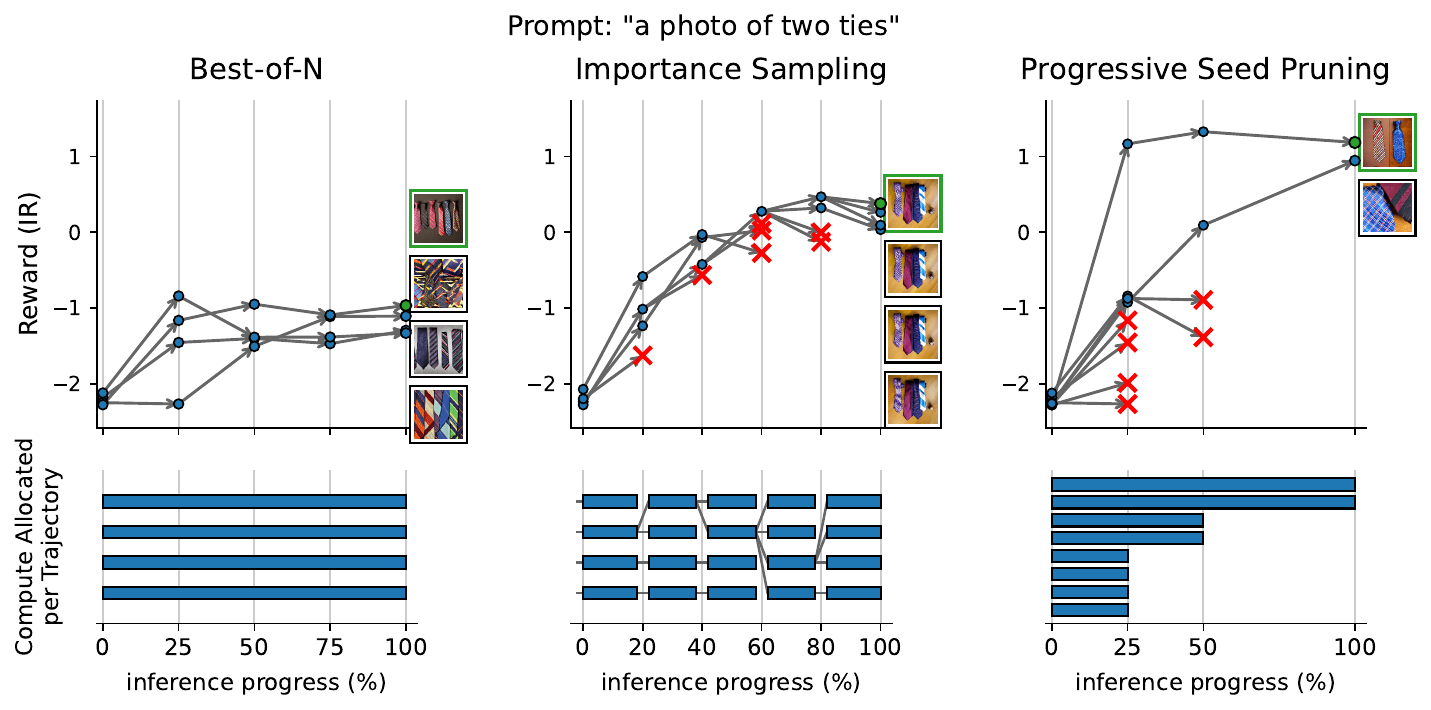}
  \caption{\textbf{Compute allocation profiles for inference-time scaling.}
  \textbf{Top}: example reward trajectories (ImageReward) over denoising progress for multiple candidates.
  \textbf{Bottom}: compute allocated per candidate as a function of progress.
  \textbf{Left:} \BoN \ runs $N$ full trajectories and selects the best final reward.
  \textbf{Middle:} importance-sampling-style methods (e.g., FK-Steering) maintain a constant particle count and periodically resample based on intermediate rewards (red crosses indicate discarded samples).
  \textbf{Right:} \PSP \ starts with more candidates and prunes low-reward trajectories early, concentrating the remaining denoising budget on a shrinking survivor set.
  In this illustration, \PSP \ considers twice as many initial seeds as the other methods under the same total number of denoising steps.}
  \label{fig:methods_trajectories}
\end{figure}

\subsection{Preliminaries}
\label{sec:prelim}

\subsubsection{Diffusion and Flow-Matching Models}
We consider conditional generative models that transform noise into a sample through a sequence of updates, conditioned on $c$ (e.g., a text prompt).

\paragraph{Diffusion models.}
In diffusion \cite{sohl-dickstein_deep_2015, ho_denoising_2020, song_scorebased_2021}, a forward noising process defines $q(x_t \mid x_0)$ for $t\in\{1,\ldots,T\}$, and a learned model parametrizes a reverse process
$p_\theta(x_{t-1}\mid x_t,t,c)$.
In common $\epsilon$-prediction parameterizations, the network predicts $\epsilon_\theta(x_t,t,c)$ and a sampler (e.g., DDIM \cite{song_denoising_2022}) uses it to produce $x_{t-1}$.
Importantly, the same network output yields a denoised estimate of the clean sample,
\begin{equation}
  \hat{x}_0(x_t,t,c) \;\approx\; \mathbb{E}[x_0 \mid x_t,t,c],
\end{equation}
which is routinely computed inside samplers and can be reused for intermediate scoring.

\paragraph{Flow-matching models.}
Flow matching models \cite{liu_flow_2022, lipman_flow_2023} learn a time-dependent velocity field $v_\theta(x,t,c)$ that defines an ODE (or SDE) transporting noise to data.
Sampling typically integrates the ODE from a high-noise state to a clean sample using a numerical solver (e.g., Euler Discrete \cite{euler_institutiones_1792, esser_scaling_2024}), yielding discrete states $\{x_t\}$ at a chosen set of noise levels.
With standard parameterizations used in practice, the same velocity prediction can be used to form a clean estimate via a one-step ``extrapolation'' to the zero-noise endpoint,
\begin{equation}
  \hat{x}_0(x_t,t,c) \;=\; x_t - \sigma_t\, v_\theta(x_t,t,c),
\end{equation}
where $\sigma_t$ denotes the current noise scale in the solver schedule.
As with diffusion, this estimate is already implicit in the solver update and can be extracted at negligible additional model cost.

\subsubsection{Setup and Assumptions}
We assume a conditional generator like diffusion or flow mathing $p_\theta$ that produces a sample by iterating discrete transitions
$p_\theta(x_{t-1}\mid x_t,t,c)$ for $t\in\{T,\ldots,1\}$ from a known noise distribution.
(Continuous-time solvers are handled by discretizing and applying pruning at discrete solver steps.)
We also assume a black-box reward function $r(x,c)$ defined on \emph{clean} samples (e.g., an aesthetic or preference model).
Our goal is to maximize reward under a fixed inference compute budget.

\subsubsection{Intermediate Reward Estimates}
\PSP \ relies on a cheap intermediate proxy for the final sample.
At each step, we compute a denoised estimate $\hat{x}_0(x_t,t,c)$ and score it:
\begin{equation}
  s_t \;=\; r(\hat{x}_0(x_t,t,c),c).
\end{equation}
Crucially, computing $\hat{x}_0$ reuses quantities already produced by the generator (noise prediction or velocity prediction), so \PSP \ does not require extra generator forward passes beyond those used for denoising. Examples of these denoised estimates $\hat{x}_0$ at different timesteps and across backbones can be found in \cref{sec:sup_x0_examples}.

\subsection{Progressive Seed Pruning (\PSP)}
\label{sec:pps}

\begin{algorithm}[tb]
\caption{Progressive Seed Pruning (\PSP)}
\label{alg:fixed_search}
\begin{algorithmic}
\Statex \textbf{Input:} Generator $p_\theta$, reward $r(\cdot,c)$, condition $c$, steps $T$, schedule $\{k_t\}_{t=0}^{T}$ with $k_T$ initial candidates and $k_0$ final survivors.
\Statex \textbf{Return:} Final survivors $\{x_0^{i}\}_{i=1}^{k_0}$
\State Sample $x_T^{i} \sim \mathcal{N}(0,I)$ for $i \in [k_T]$
\State $S \leftarrow \{x_T^{i}\}_{i=1}^{k_T}$
\For{$t \in \{T,\dots,1\}$}
  \State \textbf{Predict clean:} $\hat{x}_0^{i,t} \leftarrow \hat{x}_0(x_t^{i},t,c)$ for each $x_t^{i}\in S$
  \State \textbf{Score:} $s_{t}^{i} \leftarrow r(\hat{x}_0^{i,t},c)$
  \State \textbf{Prune:} $S \leftarrow \text{TopK}\big(S,\, k_{t-1}\big)$ by scores $\{s_t^i\}$
  \State \textbf{Denoise:} sample $x_{t-1}^{i} \sim p_\theta(x_{t-1}\mid x_t^{i},t,c)$ for each $x_t^{i}\in S$
  \State \textbf{Update:} $S \leftarrow \{x_{t-1}^{i}\}$
\EndFor
\State \textbf{Output:} return $S$
\end{algorithmic}
\end{algorithm}

\cref{alg:fixed_search} describes \PSP.
The method begins from a large pool of $k_T$ i.i.d.\ noise seeds and maintains a survivor set $S$ of partial trajectories.
At each step, \PSP:
(i) forms a denoised estimate $\hat{x}_0$ for each survivor;
(ii) scores survivors using a black-box reward on $\hat{x}_0$;
(iii) prunes to the top $k_{t-1}$ survivors; and
(iv) advances only survivors to the next step.
By pruning early, \PSP \ spends more of its total budget on evaluating \emph{more} initial seeds, while still completing full denoising for the most promising trajectories.

\paragraph{Pruning intervals.}
In practice, we prune only at a small set of predetermined steps and keep the batch size constant between prune points.
This is motivated by (a) smoothness of intermediate reward rankings across adjacent steps, and (b) systems considerations: fixed prune points with fixed survivor counts yield predictable memory usage and runtime per interval, simplifying scheduling in distributed inference.

\subsection{Compute Budget and the Effective Multiplier}
\label{sec:budget}
We measure inference compute in \emph{denoising steps}.
If an inference procedure propagates $k_t$ concurrent trajectories at step $t$, the total number of generator updates is
\begin{equation}
  C \;=\; \sum_{t=1}^{T} k_t.
\end{equation}
A standard single-sample inference run has $C=T$.
We therefore report the effective compute multiplier $\bar{N} = C/T$, which means that an inference procedure has the same compute as sampling $\bar{N}$ times from the regular sampler.

During inference we must both advance trajectories and compute intermediate scores.
For diffusion and flow-matching models, the generator forward pass already produces the quantities needed to compute $\hat{x}_0$, so intermediate scoring does not add additional generator evaluations. Furthermore, we achieve our results with guidance from ImageReward\cite{xu_imagereward_2023}, a lightweight model already used as guidance by FK-Steering \cite{singhal_general_2025} that adds minimal overhead to diffusion inference (see \cref{tab:overhead}.

\section{Experiments}
\label{sec:experiments}

\subsection{Backbone Models}
\label{sec:models}
We evaluate on Stable Diffusion v1.5 \cite{rombach_highresolution_2022} and Stable Diffusion XL \cite{podell_sdxl_2023} as diffusion models, and Stable Diffusion 3.5 (Large) \cite{esser_scaling_2024} as a flow matching model. We use widely available implementations from HuggingFace with $T=64$ steps for SD~v1.5/SDXL and $T=32$ solver steps for SD~3.5.

For \BoN \ and \PSP, we use deterministic solvers: DDIM \cite{song_denoising_2022} with $\eta=0$ for diffusion models and Euler Discrete \cite{esser_scaling_2024, euler_institutiones_1792} for SD~3.5.
Thus, all randomness arises from the initial noise seed, matching the intended use of \PSP \ as seed search.

FK-Steering relies on stochastic trajectories: in a deterministic setting, resampling generates identical children and wastes compute. For SD~v1.5 and SDXL, we follow FK-Steering authors and use DDIM with $\eta=1$ to inject noise \cite{singhal_general_2025}. For SD~3.5, however, strong stochasticity can noticeably degrade sampling quality under rectified-flow training, and the default stochastic setting in the public implementation yields poor FK-Steering performance. We therefore use a controlled-noise solver (\cref{sec:stochastic-sd35}) that injects Gaussian perturbations with a tunable scale, and we set this scale to the largest value that does not hurt baseline (non-resampled) generation quality.

\subsection{Baselines and Main Comparison}
\label{sec:baselines}

\begin{table}[h!]
  \caption{\textbf{Main results under matched compute.}
  We compare standard sampling ($\bar{N}=1$), \BoN, our main importance sampling (FK-Steering) and tree-search (DSearch) baselines, other relevant methods for inference-time compute scaling in diffusion (Noise Trajectory Search (NTS), Rollout Budget Forcing (RBF), Breadth-First Search (BFS), SVDD), and \PSP. All results are from our experiments using the public implementation of those methods and selecting parameters to match computational budget and same reward guidance: ImageReward. We report the guidance reward (ImageReward), \HPS, GenEval, and human evaluation scores on final images generated from GenEval prompts. \PPS \ achieves the best results in prompt-alignment as measured by automated scores (GenEval) and human evaluation}
  \label{tab:main_models_samplers}
  \centering
  \setlength{\tabcolsep}{4pt}

  \begin{tabular}{@{}llcccccc@{}}
    \toprule
    Model & Sampler & $T$ & $\bar{N}$ & IR $\uparrow$ & HPS $\uparrow$ & GenEval $\uparrow$ & \makecell{Human $\uparrow$} \\
    \midrule

    SD v1.5 & Standard                                   & 64 & 1 & -0.159 & 0.257 & 0.434 & 0.431 \\
    SD v1.5 & Best-of-N                                  & 64 & 4 &  0.655 & 0.273 & 0.542 & 0.594 \\
    SD v1.5 & FK-Steering \cite{singhal_general_2025}    & 64 & 4 &  0.640 & 0.261 & 0.531 & 0.569 \\
    SD v1.5 & DSearch \cite{li_dynamic_2025}             & 64 & 4 &  0.783 & 0.275 & 0.506 & 0.514 \\
    SD v1.5 & NTS \cite{ramesh_testtime_2025}            & 64 & 4 &  0.485 & 0.272 & 0.478 & -     \\
    SD v1.5 & RBF \cite{kim_inferencetime_2025}          & 64 & 4 &  0.670 & 0.274 & 0.520 & -     \\
    SD v1.5 & BFS \cite{zhang_inferencetime_2025}        & 64 & 4 &  0.820 & 0.263 & 0.564 & -     \\
    SD v1.5 & SVDD \cite{li_derivativefree_2024}    & 64 & 4 &  0.731 & 0.272 & 0.489 & -     \\
    \midrule
    SD v1.5 & \PSP                                       & 64 & 4 &  \textbf{0.827} & \textbf{0.278} & \textbf{0.574} & \textbf{0.624} \\
    \midrule\midrule

    SDXL & Standard                                      & 64 & 1 & 0.431 & 0.275 & 0.529 & 0.531 \\
    SDXL & Best-of-N                                     & 64 & 4 & 1.098 & 0.290 & 0.629 & 0.682 \\
    SDXL & FK-Steering \cite{singhal_general_2025}       & 64 & 4 & 1.189 & 0.284 & 0.627 & 0.676 \\
    SDXL & DSearch \cite{li_dynamic_2025}                & 64 & 4 & 1.186 & 0.301 & 0.589 & 0.649 \\
    SDXL & NTS \cite{ramesh_testtime_2025}               & 64 & 4 & 0.967 & 0.298 & 0.580 & -     \\
    SDXL & RBF \cite{kim_inferencetime_2025}             & 64 & 4 & 1.133 & \textbf{0.302} & 0.618 & -     \\
    SDXL & BFS \cite{zhang_inferencetime_2025}           & 64 & 4 & \textbf{1.247} & 0.285 & 0.636 & -     \\
    SDXL & SVDD \cite{li_derivativefree_2024}       & 64 & 4 & 0.682 & 0.287 & 0.556 & -     \\
    \midrule
    SDXL & \PSP                                          & 64 & 4 & 1.224 & 0.294 & \textbf{0.645} & \textbf{0.713} \\
    \midrule\midrule

    SD 3.5 & Standard                                    & 32 & 1 & 1.045 & 0.297 & 0.713 & 0.787 \\
    SD 3.5 & Best-of-N                                   & 32 & 4 & 1.336 & 0.304 & \textbf{0.747} & 0.831 \\
    SD 3.5 & FK-Steering \cite{singhal_general_2025}     & 32 & 4 & 1.294 & 0.284 & 0.742 & 0.837 \\
    SD 3.5 & DSearch \cite{li_dynamic_2025}              & 32 & 4 & 1.144 & 0.297 & 0.704 & 0.824 \\
    SD 3.5 & NTS \cite{ramesh_testtime_2025}             & 32 & 4 & 1.086 & 0.295 & 0.699 & -     \\
    SD 3.5 & RBF \cite{kim_inferencetime_2025}           & 32 & 4 & 1.253 & 0.296 & 0.738 & -     \\
    SD 3.5 & BFS \cite{zhang_inferencetime_2025}         & 32 & 4 & 1.343 & 0.285 & \textbf{0.747} & -     \\
    SD 3.5 & SVDD \cite{li_derivativefree_2024}     & 32 & 4 & 1.113 & 0.294 & 0.719 & -     \\
    \midrule
    SD 3.5 & \PSP                                        & 32 & 4 & \textbf{1.380} & \textbf{0.306} & \textbf{0.747} & \textbf{0.841} \\
    \bottomrule
  \end{tabular}
\end{table}

\cref{tab:main_models_samplers} compares gradient-free inference-time scaling strategies under matched compute. We evaluate on GenEval prompts \cite{ghosh_geneval_2023} and report: (i) the guidance reward (ImageReward \cite{xu_imagereward_2023}) of the selected output, (ii) \HPS v2 \cite{wu_human_2023} as an automated metric for human preference (iii) GenEval \cite{ghosh_geneval_2023} as a reward-agnostic measure of prompt alignment using detector-based checks, and (iv) human evaluation of prompt-alignment. Results are averaged over 3 random seeds.

\BoN \ runs $\bar{N}$ independent full trajectories and selects the highest-reward result (\cref{fig:methods_trajectories}, left).
As an importance sampling baseline, we use FK-Steering \cite{singhal_general_2025} (\cref{fig:methods_trajectories}, middle) with the authors' recommended hyperparameters ($\lambda=10$, $K=4$) and resampling schedule (20\%, 40\%, 60\%, 80\% and last step of inference progress). As a tree-search baseline, we use DSearch \cite{li_dynamic_2025} with beam and tree width set to 2 to match the compute of other methods. We also ran experiments with other inference-time compute-scaling methods under a matched computational budget (Noise Trajectory Search (NTS) \cite{ramesh_testtime_2025}, Rollover Budget Forcing (RBF) \cite{kim_inferencetime_2025}, Breadth-First Search (BFS) \cite{zhang_inferencetime_2025}, and SVDD \cite{li_derivativefree_2024}) but, given the cost of human studies, we report only automated metrics for these and reserve human evaluation for the main comparisons. For \PSP, we use a simple fixed schedule (\cref{fig:methods_trajectories}, right): start with $2\bar{N}$ seeds, prune to $\bar{N}$ at 25\% progress, and prune again to $\bar{N}/2$ at 50\% progress.

Across all backbones, \PSP \ improves over standard inference and outperforms all other methods at the same compute in prompt alignment, as measured by GenEval score and human evaluation. Furthermore, although BFS \cite{zhang_inferencetime_2025} outperforms \PSP in reward maximization with the SDXL backbone, \PSP \ still outperforms BFS and all methods in GenEval and human scores. This shows that \PSP \ is less susceptible to reward hacking \cite{skalse_defining_2025}, since it focuses on seed selection but does not interfere during the inference process with resampling aimed at increasing reward. Supplemental experiments using HPS for reward guidance are available in \cref{tab:hps_guidance}, but IR generally leads to better generalization to GenEval, making it the best reward guidance to use in practice.

The SD~3.5 results also highlight a practical weakness of resampling methods: they require stochasticity to generate diverse children, but stochasticity can degrade flow matching generation, with \BoN \ performing better than FK-Steering.
In contrast, \PSP \ remains effective with deterministic solvers because it allocates compute to seed exploration and early elimination, not to stochastic branching.

\subsection{Human Evaluation}
\label{sec:human_eval}

The human evaluation results in \cref{tab:main_models_samplers} were obtained from online annotators from Prolific with prior AI-evaluation experience. They answered the question ``Does the image align with the prompt?'' on the images generated from GenEval prompts that were also evaluated with the automated metrics, but a single seed was used per prompt/method/backbone triplet. To our knowledge, this is the first human evaluation of these inference time scaling methods for text-to-image tasks. In total, 249 annotators evaluated 8,295 images (553 per method/backbone pair, the number of GenEval prompts). Each image got 3 evaluations (24,885 ratings total) and majority vote was used to determine if an individual image aligned with its prompt. Annotator agreement was 80.3\%. \PSP \ was the best evaluated method on all three backbones. 

We closesely followed the protocol of \cite{kamath_geneval_2025} (see \cref{sec:sup_human} for our study+ details), in which the authors use a similar study to assess differences in human and GenEval evaluations. In their study, they use only the standard samplers and the scores they obtained for them on backbones shared across studies closely match (SDXL: 0.531 vs.\ 0.566; SD~3.5: 0.787 vs.\ 0.770), reproducing their results and thus validating the protocol.

\subsection{Scaling with Compute}
\label{sec:scaling}

\begin{figure}[t]
  \centering
  \newcommand{\scalewidth}{0.95\columnwidth}
  \includegraphics[width=\scalewidth]{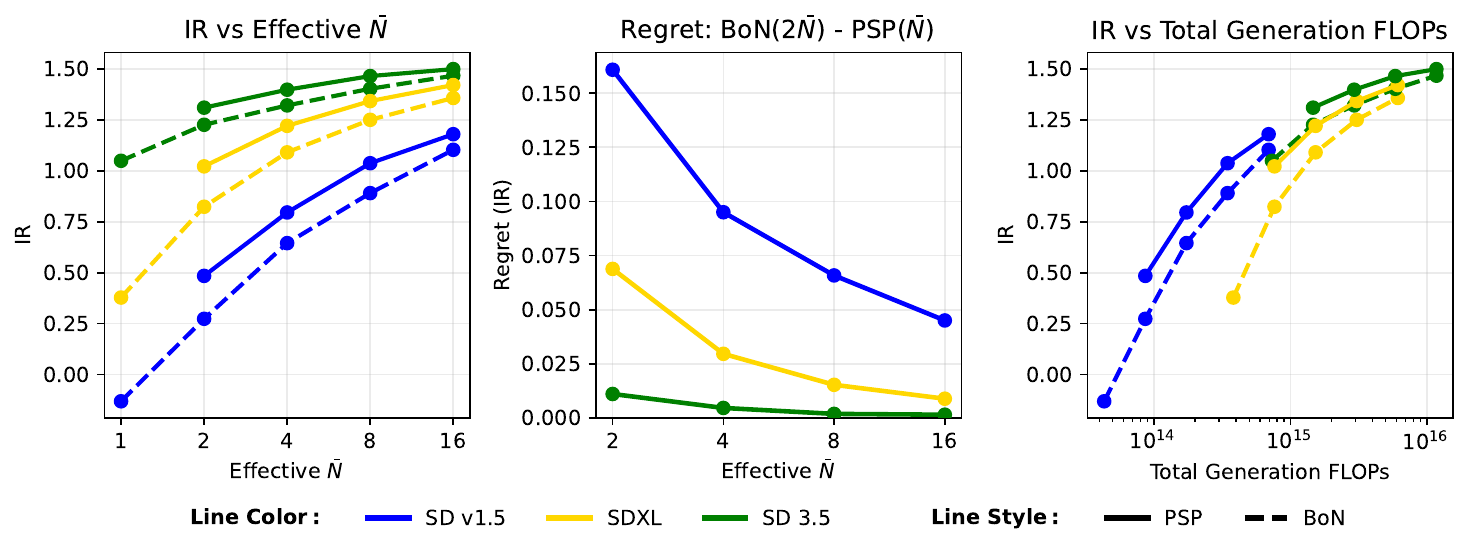}
  \caption{\textbf{Scaling behavior of \PSP.}
  \textbf{Left:} final guidance reward versus compute multiplier $\bar{N}$, comparing \PSP \ to \BoN at matched compute.
  \textbf{Middle:} \emph{regret} of \PSP \ relative to \BoN \ with double compute that would select the best seed among all $2\bar{N}$ candidates after full denoising; regret decreases with $\bar{N}$, indicating pruning becomes increasingly safe at larger budgets.
  \textbf{Right:} reward versus approximate FLOPs, showing that inference-time scaling can allow smaller models with larger $\bar{N}$ to match or surpass larger backbones under similar compute.}
  \label{fig:scaling_strategy}
\end{figure}

To test how effectively \PSP \ uses additional compute, we scale $\bar{N}$ up to 16 by doubling the number of initial seeds and survivor counts while keeping the same fractional prune points. \cref{fig:scaling_strategy} (left) shows that \PSP \ continues to improve with increasing $\bar{N}$ and consistently outperforms \BoN \ at matched compute.

Because \PSP \ starts from $2\bar{N}$ deterministic trajectories, its best possible output is upper-bounded by the best fully-denoised sample among those $2\bar{N}$ seeds.
Pruning introduces \emph{regret} relative to this upper bound when the final best seed is mistakenly discarded early. \cref{fig:scaling_strategy} (middle) shows that this regret decreases as compute increases, indicating that intermediate rankings become sufficiently reliable for pruning and that \PSP \ approaches the $2\bar{N}$-seed upper bound using only the compute of $\bar{N}$ full trajectories.

Finally, \cref{fig:scaling_strategy} (right) illustrates that inference-time scaling can change the model-compute tradeoff: smaller backbones with larger $\bar{N}$ can outperform larger models at comparable total FLOPs, highlighting \PSP \ as a practical knob for deployment-time quality scaling.

\subsection{Computational Overhead}
\label{sec:overhead}

\begin{table}[b]
  \caption{\textbf{Computational overhead of \PSP \ over \BoN.}}
  \label{tab:overhead}
  \centering
  \setlength{\tabcolsep}{5pt}

  \begin{tabular}{@{}lcccccc@{}}
    \toprule
    \multirow{2}{*}{Model}
      & \multicolumn{3}{c}{Runtime (s)}
      & \multicolumn{3}{c}{Peak VRAM (GiB)} \\
    \cmidrule(lr){2-4}
    \cmidrule(lr){5-7}
      & \makecell{\BoN \\ $(\bar N = 4)$}
      & \makecell{\PSP \\ $(\bar N = 4)$}
      & \makecell{\BoN \\ $(\bar N = 8)$}
      & \makecell{\BoN \\ $(\bar N = 4)$}
      & \makecell{\PSP \\ $(\bar N = 4)$}
      & \makecell{\BoN \\ $(\bar N = 8)$} \\
    \midrule

    SD v1.5 & 2.65  & 2.99  & 4.70  & 4.29  & 5.30  & 5.31  \\
    SDXL    & 12.48 & 13.74 & 23.90 & 8.28  & 13.27 & 10.79 \\
    SD 3.5  & 35.97 & 37.59 & 71.71 & 29.85 & 34.85 & 33.86 \\

    \bottomrule
  \end{tabular}
\end{table}

\PSP \ adds computational overhead over BoN because its intermediate samples must be decoded by the VAE and the intermediate samples must be scored by the reward model. We quantify this overhead on a single H200 in \cref{tab:overhead}, comparing \PSP, equivalent BoN ($N{=}4$), and double compute BoN($N{=}8$) under the same setup as \cref{tab:main_models_samplers}. Across all backbones, \PSP \ adds under little runtime over BoN ($N{=}4$) and is close to $2\times$ faster than BoN ($N{=}8$). Relative overhead also diminishes with model size, as each diffusion step becomes proportionally more expensive. Regarding Peak VRAM usage, the built-in VAE slicing option, which decodes images sequentially through the VAE instead of in parallel. This was used in \cref{tab:overhead}, adding less than $0.15\,s$ of runtime and bringing \PSP's peak VRAM close to BoN($N{=}4$) on SD~v1.5 and SD~3.5. SDXL is the one case where it remains noticeably higher, because its VAE is proportionally larger. In effect, \PSP($\bar N{=}4$) considers a BoN($N{=}8$)-sized seed pool at near-BoN($N{=}4$) cost.

\subsection{Tuning \PSP \ Schedule}
\label{sec:tuning}

\begin{figure*}[t]
  \centering
  \includegraphics[width=\textwidth]{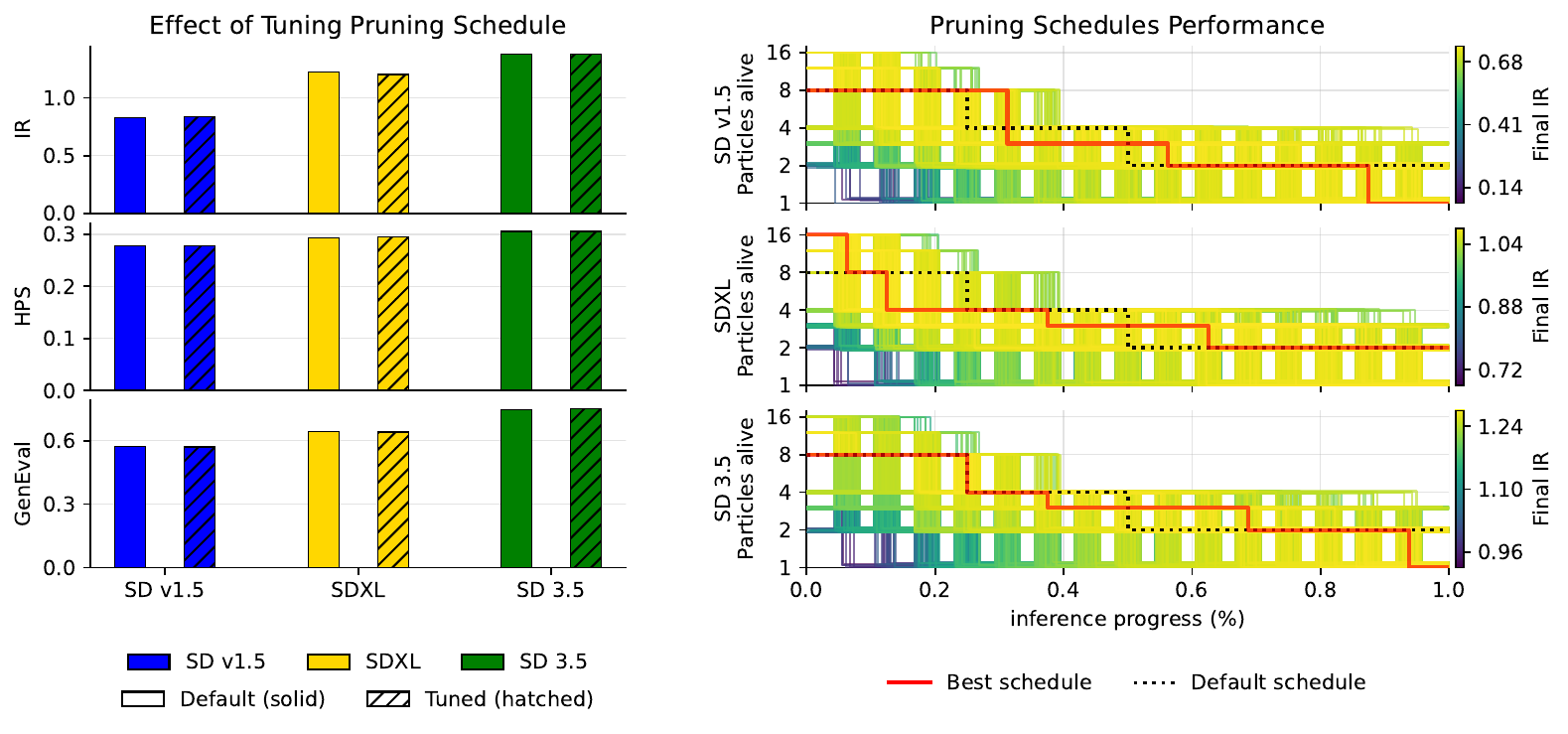}
  \caption{\textbf{Tuning the pruning strategy improves \PSP.}
  {\bf Left.} We compare our default, off-the-shelf schedule (solid) to the best schedule found by grid search under the same compute budget (hatched). We search on prompts from the IR Benchmark and report results on prompts from GenEval, using IR as guidance signal. {\bf Right.} Visualization of all schedules in the grid search colored by their performance on Benchmark IR. Better schedules above and jitter added for better visualization, with best and default schedules highlighted.}
  \label{fig:tuning_comparison}
\end{figure*}

So far we used a simple default schedule motivated by compute matching. For large enough $T$, since $\sum_{i=0}^k\frac{1}{2^i}=2-\frac{1}{2^k}$, a geometric halving schedule lets us evaluate an arbitrarily large number of initial seeds with effective cost multiplier $\bar N=2$. Scaling the schedule by a factor $m$ (multiplying the number of particles at every stage by $m$) yields $\bar N = 2m$. Concretely, there is a tradeoff as larger $k$ requires pruning to start earlier at less informative stages, so our default schedule (\cref{fig:methods_trajectories}, right) has two pruning points, which allows considering $2\bar N$ initial seeds with compute multiplier $\bar N$ (double the seeds of \BoN), and uses $m=2$ (thus $\bar N=4$), matching compute settings used in \cite{singhal_general_2025}. This choice makes \PSP\ immediately usable off-the-shelf and enables a fair baseline comparison in \cref{tab:main_models_samplers} without hyperparameter tuning.

However, \PSP \ has a rich design space: the number of initial candidates, the timing of prune points, and the survivor counts can all be varied while maintaining the same total budget. We therefore perform a grid search over schedules that preserve $\bar{N}=4$ for each model and reward choice, select schedules that maximize reward on a held-out prompt set (Benchmark \IR; details in \cref{sec:gridsearch}), and test on prompts from GenEval. The left panel in \cref{fig:tuning_comparison} shows that tuned schedules show little to no improvement over our default schedule on guidance rewards, HPS, and GenEval scores for this task. The right panel shows all schedules searched colored by their performance on IR benchmark, highlighting the optimal chosen schedule and the default schedule we used before without tuning. We can see that they are not dramatically different, so our default halving strategy can be considered a good off-the-shelf configuration. Differences mostly lie in that tuned strategies tend to give up one sample close to the end in order to develop 3 samples beyond 50\% of the inference process.

A key practical advantage is that when sampling is deterministic, this tuning can be done cheaply. We precompute trajectories and intermediate rewards for each prompt across a pool of initial seeds. Because the outcome of \PSP \ is fully determined by the schedule and the precomputed trajectories, we can simulate many schedules without rerunning the generator, enabling rapid per-task tuning in domains where the default schedule might be far from the optimal one.

\cref{fig:reward_distributions_and_binned_geneval} (left) shows positive correlation between tuning (Benchmark IR) and test set (GenEval) performance across schedules, indicating that better schedules tend to generalize across prompt sets. For our default schedule, the distribution of final IR is also very similar across the two datasets (\cref{fig:reward_distributions_and_binned_geneval}, middle). \cref{fig:reward_distributions_and_binned_geneval} (right) shows diminishing returns of IR on GenEval: beyond a high-IR regime, further IR gains translate weakly to GenEval, explaining the smaller GenEval improvements for strong backbones such as SD~3.5. \PSP\ optimizes whatever guidance signal is provided, and can directly benefit from stronger reward models as they become available.

\subsection{Comparison to Finetuned Models}
\label{sec:finetuned}

\begin{table}[tb]
  \caption{\textbf{\PSP \ complements reward-based finetuning.}
  \PSP\ on a non-finetuned model exceeds standard sampling on a DPO-finetuned model, and applying \PSP\ to a DPO-finetuned model yields further gains over \BoN.}
  \label{tab:dpo_pps_ir_geneval}
  \centering
  \setlength{\tabcolsep}{4pt}

  \newcommand{\dpoyes}{\textcolor{green!50!black}{\large$\checkmark$}}
  \newcommand{\dpono}{\textcolor{red!70!black}{\large$\times$}}

  \begin{tabular}{@{}lclcccc@{}}
    \toprule
    Model & DPO & Sampler & $T$ & $N$ & IR $\uparrow$ & GenEval $\uparrow$ \\
    \midrule

    SD v1.5 & \dpoyes & Standard       & 64 & 1 & -0.022  & 0.454 \\
    SD v1.5 & \dpoyes & Best-of-N      & 64 & 4 & 0.751   & 0.562 \\
    \midrule
    SD v1.5 & \dpono  & \PSP            & 64 & 4 & 0.827   & 0.574 \\
    SD v1.5 & \dpoyes & \PSP            & 64 & 4 & {\bf 0.907}  & {\bf 0.593} \\
    \midrule\midrule

    SDXL    & \dpoyes & Standard       & 64 & 1 & 0.894   & 0.581 \\
    SDXL    & \dpoyes & Best-of-N      & 64 & 4 & 1.300   & 0.657 \\
    \midrule
    SDXL    & \dpono  & \PSP            & 64 & 4 & 1.224   & 0.645 \\
    SDXL    & \dpoyes & \PSP            & 64 & 4 & {\bf 1.365}  & {\bf 0.667} \\
    \bottomrule
  \end{tabular}
\end{table}

Reward models can also be used to \emph{finetune} generators, amortizing reward optimization across many future inference calls.
We compare against DPO-finetuned backbones \cite{wallace_diffusion_2023} trained on the Pick-a-Pic dataset \cite{kirstain_pickapic_2023}.
\cref{tab:dpo_pps_ir_geneval} shows two takeaways.
First, \PSP \ applied to a non finetuned backbone with a modest compute multiplier outperforms standard inference on a finetuned backbone.
Second, \PSP \ and finetuning are complementary: applying \PSP \ on a finetuned backbone yields further improvements and outperforms \BoN \ under similar compute.
Thus, whenever additional inference-time compute is available, \PSP \ remains beneficial even in the presence of finetuned models.

\subsection{\PSP \ over Multiple Prompts}
\label{sec:multiprompts}

Although our primary focus is seed search, \PSP \ applies to any discrete pool of candidates available at inference time.
A practical example is prompt selection: modern systems often rewrite user prompts \cite{betker_improving_, esser_scaling_2024, deng_emerging_2025} to expand it and add detail, and different rewrites can substantially change the quality of the outcome. Quantitatively, analysis from previous work \cite{kamath_geneval_2025} has shown that prompt rewriting can bring large gains in GenEval score.
We therefore apply \PSP \ to select among multiple prompt rewrites for a \emph{fixed} noise seed.
\cref{tab:pps_multiprompts} shows that \PSP \ outperforms \BoN \ at equal compute in this setting.

In our implementation, we generate prompt rewrites once per prompt with ChatGPT 5.2 \cite{singh_openai_2025} (details in \cref{sec:rewriting}) and treat them as the initial candidate set.
This highlights another practical advantage of \PSP \ over resampling-based methods: importance sampling over prompts would require developing a stochastic ``prompt mutation'' kernel that generates children prompts repeatedly during inference and also incur the additional cost of sampling from a LLM multiple times, whereas \PSP \ only requires a fixed candidate set once per inference.

\begin{table}[t]
  \caption{\textbf{Searching over different prompts with fixed initial seeds.}
  \PSP \ can optimize over any discrete set of initial conditions, including a set of prompts.}
  \label{tab:pps_multiprompts}
  \centering
  \setlength{\tabcolsep}{4pt}

  \begin{tabular}{@{}llcccc@{}}
    \toprule
    Model & Sampler & $T$ & $N$ & IR $\uparrow$ & GenEval $\uparrow$ \\
    \midrule
    SD v1.5  & Standard     & 64 & 1 & -0.230 & 0.391 \\
    SD v1.5  & Best-of-N    & 64 & 4 & 0.641  & 0.512 \\
    \midrule
    SD v1.5  & \PSP          & 64 & 4 & {\bf 0.782} & {\bf 0.533} \\
    \midrule\midrule
    SDXL     & Standard     & 64 & 1 & 0.471  & 0.501 \\
    SDXL     & Best-of-N    & 64 & 4 & 1.256  & 0.598 \\
    \midrule
    SDXL     & \PSP          & 64 & 4 & {\bf 1.319} & {\bf 0.604} \\
    \bottomrule
  \end{tabular}
\end{table}

\section{Discussion}
\label{sec:conclusion}

We study inference-time scaling for diffusion and flow-matching models under black-box rewards. Our main contribution is showing that if we relax the constraint of constant-memory during inference, as is natural in the distributed systems that host most generative models today, then even a very simple early-pruning strategy can be more compute-efficient than stronger baselines. Across diffusion and a flow-matching backbones, \PSP\ consistently improves reward-guided selection and achieves higher GenEval scores than \BoN\, tree-search, and importance-sampling methods at matched compute, and continues to scale as compute increases.

A practical implication of this design is that \PSP\ does not require stochasticity. Resampling-based approaches rely on stochastic samplers to generate diverse children from the same intermediate state. Without it, resampling collapses to repeated children and that offers no benefit over \BoN. In contrast, \PSP\ uses compute to consider more initial seeds and increasing its chances of retaining a good one. Our strong results suggest that in some tasks this is a more important and compute-efficient lever for diffusion inference than repeatedly ``improving'' the current best candidates via resampling. This determinism is also well aligned with the growing adoption of flow-matching over diffusion, where deterministic samplers are the standard. Moreover, because \PSP\ is deterministic given the initial seeds, intermediate scores can be precomputed once, enabling efficient offline search over pruning schedules tailored to a specific task, reward, and prompt distribution without rerunning the generative model for each candidate schedule.

\PSP\ has two main limitations. First, it presumes a scalar reward, so multi-objective or hard-to-quantify settings (e.g.\ ControlNet-style \cite{zhang_adding_2023} constrained generation, where the signal must jointly balance text alignment with strict spatial consistency) are less natural to express as the single ranking \PSP\ prunes on. In practice this is often surmountable: many strict-spatial tasks admit scalar perceptual losses (e.g.\ IoU, OKS) that \PSP\ can rank on directly, and at matched compute it could still outperform other search/resampling methods. The deeper, shared limitation of such selection methods is the lack of \emph{directional} guidance in space, which alternatives supply only with tradeoffs: ControlNet adds training, while gradient-based guidance requires a differentiable reward and far larger inference cost for backpropagation. Second, \PSP's advantage stems from \emph{when} its target features are decided. For prompt alignment, the salient features are coarse and fixed early in sampling, so outcomes hinge on the initial seed and front-loading a larger seed pool pays off. For objectives dominated by fine-grained detail generated late in the trajectory, such as aesthetic quality, early seed selection matters less, and the gains shrink accordingly. This is visible in the comparatively lower HPS (which also reflects aesthetics) for SDXL in \cref{tab:main_models_samplers}.

Our novelty is the finding that maintaining \textbf{variable particle counts} during inference is a strong, underexplored inference-time scaling knob for diffusion and flow-matching generators in settings where the \emph{constant-memory} constraint is not binding, as is typical in production deployments, where inference runs on multi-GPU servers and memory is allocated elastically. Empirically, this allocation consistently beats stronger constant-memory baselines at matched generator compute and enables offline schedule tuning under deterministic sampling. Taken together, our results provide a simple but powerful design principle for inference-time scaling in modern generative models.

\section*{Acknowledgments}
\label{sec:ack}

This study was supported by the Technology Innovation Institute (TII) through the project ``Endowing AI Agents with Hierarchical Compositional \& Spatial Reasoning.''

%
%
\bibliographystyle{splncs04}
\bibliography{references}

@inproceedings{sohl-dickstein_deep_2015,
  title={Deep unsupervised learning using nonequilibrium thermodynamics},
  author={Sohl-Dickstein, Jascha and Weiss, Eric and Maheswaranathan, Niru and Ganguli, Surya},
  booktitle={International conference on machine learning},
  pages={2256--2265},
  year={2015},
  organization={pmlr}
}

@article{ho_denoising_2020,
  title={Denoising diffusion probabilistic models},
  author={Ho, Jonathan and Jain, Ajay and Abbeel, Pieter},
  journal={Advances in neural information processing systems},
  volume={33},
  pages={6840--6851},
  year={2020}
}

@inproceedings{song_scorebased_2021,
  author       = {Yang Song and
                  Jascha Sohl{-}Dickstein and
                  Diederik P. Kingma and
                  Abhishek Kumar and
                  Stefano Ermon and
                  Ben Poole},
  title        = {Score-Based Generative Modeling through Stochastic Differential Equations},
  booktitle    = {9th International Conference on Learning Representations, {ICLR} 2021,
                  Virtual Event, Austria, May 3-7, 2021},
  publisher    = {OpenReview.net},
  year         = {2021},
  url          = {https://openreview.net/forum?id=PxTIG12RRHS},
  bibsource    = {dblp computer science bibliography, https://dblp.org}
}

@inproceedings{liu_flow_2022,
  author       = {Xingchao Liu and
                  Chengyue Gong and
                  Qiang Liu},
  title        = {Flow Straight and Fast: Learning to Generate and Transfer Data with
                  Rectified Flow},
  booktitle    = {The Eleventh International Conference on Learning Representations,
                  {ICLR} 2023, Kigali, Rwanda, May 1-5, 2023},
  publisher    = {OpenReview.net},
  year         = {2023},
  url          = {https://openreview.net/forum?id=XVjTT1nw5z},
  bibsource    = {dblp computer science bibliography, https://dblp.org}
}

@inproceedings{lipman_flow_2023,
  author       = {Yaron Lipman and
                  Ricky T. Q. Chen and
                  Heli Ben{-}Hamu and
                  Maximilian Nickel and
                  Matthew Le},
  title        = {Flow Matching for Generative Modeling},
  booktitle    = {The Eleventh International Conference on Learning Representations,
                  {ICLR} 2023, Kigali, Rwanda, May 1-5, 2023},
  publisher    = {OpenReview.net},
  year         = {2023},
  url          = {https://openreview.net/forum?id=PqvMRDCJT9t},
  bibsource    = {dblp computer science bibliography, https://dblp.org}
}

@article{yao_tree_2023,
  title={Tree of thoughts: Deliberate problem solving with large language models},
  author={Yao, Shunyu and Yu, Dian and Zhao, Jeffrey and Shafran, Izhak and Griffiths, Tom and Cao, Yuan and Narasimhan, Karthik},
  journal={Advances in neural information processing systems},
  volume={36},
  pages={11809--11822},
  year={2023}
}

@inproceedings{besta_graph_2024,
  title={Graph of thoughts: Solving elaborate problems with large language models},
  author={Besta, Maciej and Blach, Nils and Kubicek, Ales and Gerstenberger, Robert and Podstawski, Michal and Gianinazzi, Lukas and Gajda, Joanna and Lehmann, Tomasz and Niewiadomski, Hubert and Nyczyk, Piotr and others},
  booktitle={Proceedings of the AAAI conference on artificial intelligence},
  volume={38},
  number={16},
  pages={17682--17690},
  year={2024}
}

@article{cobbe_training_2021,
  author       = {Karl Cobbe and
                  Vineet Kosaraju and
                  Mohammad Bavarian and
                  Mark Chen and
                  Heewoo Jun and
                  Lukasz Kaiser and
                  Matthias Plappert and
                  Jerry Tworek and
                  Jacob Hilton and
                  Reiichiro Nakano and
                  Christopher Hesse and
                  John Schulman},
  title        = {Training Verifiers to Solve Math Word Problems},
  journal      = {CoRR},
  volume       = {abs/2110.14168},
  year         = {2021},
  url          = {https://arxiv.org/abs/2110.14168},
  eprinttype   = {arXiv},
  eprint       = {2110.14168},
  bibsource    = {dblp computer science bibliography, https://dblp.org}
}

@article{yao_webshop_2023,
  title={Webshop: Towards scalable real-world web interaction with grounded language agents},
  author={Yao, Shunyu and Chen, Howard and Yang, John and Narasimhan, Karthik},
  journal={Advances in Neural Information Processing Systems},
  volume={35},
  pages={20744--20757},
  year={2022}
}

@article{valmeekam_planning_2023,
  title={On the planning abilities of large language models-a critical investigation},
  author={Valmeekam, Karthik and Marquez, Matthew and Sreedharan, Sarath and Kambhampati, Subbarao},
  journal={Advances in neural information processing systems},
  volume={36},
  pages={75993--76005},
  year={2023}
}

@article{qi_not_2024,
  author       = {Zipeng Qi and
                  Lichen Bai and
                  Haoyi Xiong and
                  Zeke Xie},
  title        = {Not All Noises Are Created Equally:Diffusion Noise Selection and Optimization},
  journal      = {CoRR},
  volume       = {abs/2407.14041},
  year         = {2024},
  url          = {https://doi.org/10.48550/arXiv.2407.14041},
  doi          = {10.48550/ARXIV.2407.14041},
  eprinttype   = {arXiv},
  eprint       = {2407.14041},
  bibsource    = {dblp computer science bibliography, https://dblp.org}
}

@inproceedings{xu_good_2025,
  title={Good seed makes a good crop: Discovering secret seeds in text-to-image diffusion models},
  author={Xu, Katherine and Zhang, Lingzhi and Shi, Jianbo},
  booktitle={2025 IEEE/CVF Winter Conference on Applications of Computer Vision (WACV)},
  pages={3024--3034},
  year={2025},
  organization={IEEE}
}

@article{ma_inferencetime_2025,
  author       = {Nanye Ma and
                  Shangyuan Tong and
                  Haolin Jia and
                  Hexiang Hu and
                  Yu{-}Chuan Su and
                  Mingda Zhang and
                  Xuan Yang and
                  Yandong Li and
                  Tommi S. Jaakkola and
                  Xuhui Jia and
                  Saining Xie},
  title        = {Inference-Time Scaling for Diffusion Models beyond Scaling Denoising
                  Steps},
  journal      = {CoRR},
  volume       = {abs/2501.09732},
  year         = {2025},
  url          = {https://doi.org/10.48550/arXiv.2501.09732},
  doi          = {10.48550/ARXIV.2501.09732},
  eprinttype   = {arXiv},
  eprint       = {2501.09732},
  bibsource    = {dblp computer science bibliography, https://dblp.org}
}

@inproceedings{singhal_general_2025,
	series = {Proceedings of machine learning research},
	title = {A general framework for inference-time scaling and steering of diffusion models},
	volume = {267},
	url = {https://proceedings.mlr.press/v267/singhal25b.html},
	booktitle = {Proceedings of the 42nd international conference on machine learning},
	publisher = {PMLR},
	author = {Singhal, Raghav and Horvitz, Zachary and Teehan, Ryan and Ren, Mengye and Yu, Zhou and Mckeown, Kathleen and Ranganath, Rajesh},
	editor = {Singh, Aarti and Fazel, Maryam and Hsu, Daniel and Lacoste-Julien, Simon and Berkenkamp, Felix and Maharaj, Tegan and Wagstaff, Kiri and Zhu, Jerry},
	month = jul,
	year = {2025},
	pages = {55810--55827},
}

@article{li_dynamic_2025,
  author       = {Xiner Li and
                  Masatoshi Uehara and
                  Xingyu Su and
                  Gabriele Scalia and
                  Tommaso Biancalani and
                  Aviv Regev and
                  Sergey Levine and
                  Shuiwang Ji},
  title        = {Dynamic Search for Inference-Time Alignment in Diffusion Models},
  journal      = {CoRR},
  volume       = {abs/2503.02039},
  year         = {2025},
  url          = {https://doi.org/10.48550/arXiv.2503.02039},
  doi          = {10.48550/ARXIV.2503.02039},
  eprinttype   = {arXiv},
  eprint       = {2503.02039},
  bibsource    = {dblp computer science bibliography, https://dblp.org}
}

@article{zhang_inferencetime_2025,
  author       = {Xiangcheng Zhang and
                  Haowei Lin and
                  Haotian Ye and
                  James Y. Zou and
                  Jianzhu Ma and
                  Yitao Liang and
                  Yilun Du},
  title        = {Inference-time Scaling of Diffusion Models through Classical Search},
  journal      = {CoRR},
  volume       = {abs/2505.23614},
  year         = {2025},
  url          = {https://doi.org/10.48550/arXiv.2505.23614},
  doi          = {10.48550/ARXIV.2505.23614},
  eprinttype   = {arXiv},
  eprint       = {2505.23614},
  bibsource    = {dblp computer science bibliography, https://dblp.org}
}

@article{oshima_inferencetime_2025,
  title={Inference-time text-to-video alignment with diffusion latent beam search},
  author={Oshima, Yuta and Suzuki, Masahiro and Matsuo, Yutaka and Furuta, Hiroki},
  journal={Advances in Neural Information Processing Systems},
  volume={38},
  pages={13170--13216},
  year={2026}
}

@article{fox_kldsampling_2001,
	title = {{KLD}-sampling: {Adaptive} particle filters},
	volume = {14},
	journal = {Advances in neural information processing systems},
	author = {Fox, Dieter},
	year = {2001},
}

@inproceedings{chung_diffusion_2024,
  author       = {Hyungjin Chung and
                  Jeongsol Kim and
                  Michael Thompson McCann and
                  Marc Louis Klasky and
                  Jong Chul Ye},
  title        = {Diffusion Posterior Sampling for General Noisy Inverse Problems},
  booktitle    = {The Eleventh International Conference on Learning Representations,
                  {ICLR} 2023, Kigali, Rwanda, May 1-5, 2023},
  publisher    = {OpenReview.net},
  year         = {2023},
  url          = {https://openreview.net/forum?id=OnD9zGAGT0k},
  bibsource    = {dblp computer science bibliography, https://dblp.org}
}

@inproceedings{bansal_universal_2023,
  title={Universal guidance for diffusion models},
  author={Bansal, Arpit and Chu, Hong-Min and Schwarzschild, Avi and Sengupta, Roni and Goldblum, Micah and Geiping, Jonas and Goldstein, Tom},
  booktitle={International Conference on Learning Representations},
  volume={2024},
  pages={51304--51323},
  year={2024}
}

@inproceedings{tang_inferencetime_2024,
  author       = {Zhiwei Tang and
                  Jiangweizhi Peng and
                  Jiasheng Tang and
                  Mingyi Hong and
                  Fan Wang and
                  Tsung{-}Hui Chang},
  editor       = {Aarti Singh and
                  Maryam Fazel and
                  Daniel Hsu and
                  Simon Lacoste{-}Julien and
                  Felix Berkenkamp and
                  Tegan Maharaj and
                  Kiri Wagstaff and
                  Jerry Zhu},
  title        = {Inference-Time Alignment of Diffusion Models with Direct Noise Optimization},
  booktitle    = {Forty-second International Conference on Machine Learning, {ICML}
                  2025, Vancouver, BC, Canada, July 13-19, 2025},
  series       = {Proceedings of Machine Learning Research},
  volume       = {267},
  publisher    = {{PMLR} / OpenReview.net},
  year         = {2025},
  url          = {https://proceedings.mlr.press/v267/tang25h.html},
  bibsource    = {dblp computer science bibliography, https://dblp.org}
}

@inproceedings{black_training_2024,
  title={Training diffusion models with reinforcement learning},
  author={Black, Kevin and Janner, Michael and Du, Yilun and Kostrikov, Ilya and Levine, Sergey},
  booktitle={International Conference on Learning Representations},
  volume={2024},
  pages={4965--4987},
  year={2024}
}

@inproceedings{yang_using_2024,
  title={Using human feedback to fine-tune diffusion models without any reward model},
  author={Yang, Kai and Tao, Jian and Lyu, Jiafei and Ge, Chunjiang and Chen, Jiaxin and Shen, Weihan and Zhu, Xiaolong and Li, Xiu},
  booktitle={Proceedings of the IEEE/CVF Conference on Computer Vision and Pattern Recognition},
  pages={8941--8951},
  year={2024}
}

@inproceedings{wallace_diffusion_2023,
  title={Diffusion model alignment using direct preference optimization},
  author={Wallace, Bram and Dang, Meihua and Rafailov, Rafael and Zhou, Linqi and Lou, Aaron and Purushwalkam, Senthil and Ermon, Stefano and Xiong, Caiming and Joty, Shafiq and Naik, Nikhil},
  booktitle={Proceedings of the IEEE/CVF Conference on Computer Vision and Pattern Recognition},
  pages={8228--8238},
  year={2024}
}

@book{delmoral_feynmankac_2004,
	address = {New York, NY},
	series = {Probability and its {Applications}},
	title = {Feynman-{Kac} {Formulae}},
	copyright = {http://www.springer.com/tdm},
	isbn = {978-1-4419-1902-1 978-1-4684-9393-1},
	url = {http://link.springer.com/10.1007/978-1-4684-9393-1},
	doi = {10.1007/978-1-4684-9393-1},
	urldate = {2026-03-05},
	publisher = {Springer New York},
	author = {Del Moral, Pierre},
	editor = {Gani, J. and Heyde, C. C. and Kurtz, T. G.},
	year = {2004},
}

@inproceedings{guimaraes_diffusion_2026,
  author       = {Rog{\'{e}}rio Guimar{\~{a}}es and
                  Frank Xiao and
                  Pietro Perona and
                  Markus Marks},
  title        = {Diffusion-Based Action Recognition Generalizes to Untrained Domains},
  booktitle    = {{IEEE/CVF} Winter Conference on Applications of Computer Vision, {WACV}
                  2026, Tucson, AZ, USA, March 6-10, 2026},
  pages        = {5919--5933},
  publisher    = {{IEEE}},
  year         = {2026},
  url          = {https://doi.org/10.1109/WACV61042.2026.00573},
  doi          = {10.1109/WACV61042.2026.00573},
  bibsource    = {dblp computer science bibliography, https://dblp.org}
}

@inproceedings{luo_diffusion_2023,
  author       = {Grace Luo and
                  Lisa Dunlap and
                  Dong Huk Park and
                  Aleksander Holynski and
                  Trevor Darrell},
  editor       = {Alice Oh and
                  Tristan Naumann and
                  Amir Globerson and
                  Kate Saenko and
                  Moritz Hardt and
                  Sergey Levine},
  title        = {Diffusion Hyperfeatures: Searching Through Time and Space for Semantic
                  Correspondence},
  booktitle    = {Advances in Neural Information Processing Systems 36: Annual Conference
                  on Neural Information Processing Systems 2023, NeurIPS 2023, New Orleans,
                  LA, USA, December 10 - 16, 2023},
  year         = {2023},
  url          = {http://papers.nips.cc/paper\_files/paper/2023/hash/942032b61720a3fd64897efe46237c81-Abstract-Conference.html},
  bibsource    = {dblp computer science bibliography, https://dblp.org}
}

@inproceedings{park_understanding_2023,
  author       = {Yong{-}Hyun Park and
                  Mingi Kwon and
                  Jaewoong Choi and
                  Junghyo Jo and
                  Youngjung Uh},
  editor       = {Alice Oh and
                  Tristan Naumann and
                  Amir Globerson and
                  Kate Saenko and
                  Moritz Hardt and
                  Sergey Levine},
  title        = {Understanding the Latent Space of Diffusion Models through the Lens of Riemannian Geometry},
  booktitle    = {Advances in Neural Information Processing Systems 36: Annual Conference
                  on Neural Information Processing Systems 2023, NeurIPS 2023, New Orleans,
                  LA, USA, December 10 - 16, 2023},
  year         = {2023},
  url          = {http://papers.nips.cc/paper\_files/paper/2023/hash/4bfcebedf7a2967c410b64670f27f904-Abstract-Conference.html},
  bibsource    = {dblp computer science bibliography, https://dblp.org}
}

@article{carmona_interacting_2009,
	title = {Interacting particle systems for the computation of rare credit portfolio losses},
	volume = {13},
	issn = {0949-2984, 1432-1122},
	url = {http://link.springer.com/10.1007/s00780-009-0098-8},
	doi = {10.1007/s00780-009-0098-8},
	language = {en},
	number = {4},
	urldate = {2026-03-05},
	journal = {Finance and Stochastics},
	author = {Carmona, René and Fouque, Jean-Pierre and Vestal, Douglas},
	month = sep,
	year = {2009},
	pages = {613--633},
}

@article{wu_practical_2024,
  title={Practical and asymptotically exact conditional sampling in diffusion models},
  author={Wu, Luhuan and Trippe, Brian and Naesseth, Christian and Blei, David and Cunningham, John P},
  journal={Advances in Neural Information Processing Systems},
  volume={36},
  pages={31372--31403},
  year={2023}
}

@inproceedings{li_derivativefree_2024,
  author       = {Xiner Li and
                  Yulai Zhao and
                  Chenyu Wang and
                  Gabriele Scalia and
                  G{\"{o}}kcen Eraslan and
                  Surag Nair and
                  Tommaso Biancalani and
                  Shuiwang Ji and
                  Aviv Regev and
                  Sergey Levine and
                  Masatoshi Uehara},
  editor       = {Danielle Belgrave and
                  Cheng Zhang and
                  Laura N. Montoya and
                  Hsuan{-}Tien Lin and
                  Razvan Pascanu and
                  Piotr Koniusz and
                  Marzyeh Ghassemi and
                  Nancy Chen and
                  Iv{\'{a}}n Vladimir Meza Ru{\'{\i}}z and
                  Arturo Loaiza{-}Bonilla},
  title        = {Derivative-Free Guidance in Continuous and Discrete Diffusion Models
                  with Soft Value-based Decoding},
  booktitle    = {Advances in Neural Information Processing Systems 38: Annual Conference
                  on Neural Information Processing Systems 2025, NeurIPS 2025, San Diago,
                  CA, USA, December 2-7, 2025 / Mexico City, Mexico, November 30 - December
                  5, 2025},
  year         = {2025},
  url          = {http://papers.nips.cc/paper\_files/paper/2025/hash/899af0d66d8850318a20781484416152-Abstract-Conference.html},
  bibsource    = {dblp computer science bibliography, https://dblp.org}
}

@article{ramesh_testtime_2025,
  title={Test-time scaling of diffusion models via noise trajectory search},
  author={Ramesh, Vignav and Mardani, Morteza},
  journal={Advances in Neural Information Processing Systems},
  volume={38},
  pages={87284--87317},
  year={2026}
}

@misc{lee_adaptive_2025,
	title = {Adaptive {Inference}-{Time} {Scaling} via {Cyclic} {Diffusion} {Search}},
	url = {http://arxiv.org/abs/2505.14036},
	doi = {10.48550/arXiv.2505.14036},
	urldate = {2026-03-05},
	publisher = {arXiv},
	author = {Lee, Gyubin and Bao, Truong Nhat Nguyen and Yoon, Jaesik and Lee, Dongwoo and Kim, Minsu and Bengio, Yoshua and Ahn, Sungjin},
	month = oct,
	year = {2025},
	note = {arXiv:2505.14036 [cs]},
}

@article{elvira_performance_2021,
	title = {On the performance of particle filters with adaptive number of particles},
	volume = {31},
	number = {6},
	journal = {Statistics and Computing},
	publisher = {Springer},
	author = {Elvira, Vı́ctor and Miguez, Joaquı́n and Djurić, Petar M},
	year = {2021},
	pages = {81},
}

@inproceedings{karnin_almost_2013,
	address = {Atlanta, Georgia, USA},
	series = {Proceedings of machine learning research},
	title = {Almost optimal exploration in multi-armed bandits},
	volume = {28},
	url = {https://proceedings.mlr.press/v28/karnin13.html},
	booktitle = {Proceedings of the 30th international conference on machine learning},
	publisher = {PMLR},
	author = {Karnin, Zohar and Koren, Tomer and Somekh, Oren},
	editor = {Dasgupta, Sanjoy and McAllester, David},
	month = jun,
	year = {2013},
	note = {Number: 3},
	pages = {1238--1246},
}

@inproceedings{jamieson_nonstochastic_2015,
  title={Non-stochastic best arm identification and hyperparameter optimization},
  author={Jamieson, Kevin and Talwalkar, Ameet},
  booktitle={Artificial intelligence and statistics},
  pages={240--248},
  year={2016},
  organization={PMLR}
}

@article{kim_inferencetime_2025,
  title={Inference-time scaling for flow models via stochastic generation and rollover budget forcing},
  author={Kim, Jaihoon and Yoon, Taehoon and Hwang, Jisung and Sung, Minhyuk},
  journal={Advances in Neural Information Processing Systems},
  volume={38},
  pages={30830--30864},
  year={2026}
}

@inproceedings{song_denoising_2022,
  author       = {Jiaming Song and
                  Chenlin Meng and
                  Stefano Ermon},
  title        = {Denoising Diffusion Implicit Models},
  booktitle    = {9th International Conference on Learning Representations, {ICLR} 2021,
                  Virtual Event, Austria, May 3-7, 2021},
  publisher    = {OpenReview.net},
  year         = {2021},
  url          = {https://openreview.net/forum?id=St1giarCHLP},
  bibsource    = {dblp computer science bibliography, https://dblp.org}
}

@book{euler_institutiones_1792,
	title = {Institutiones calculi integralis},
	volume = {1},
	publisher = {impensis Academiae imperialis scientiarum},
	author = {Euler, Leonhard},
	year = {1792},
}

@inproceedings{esser_scaling_2024,
  title={Scaling rectified flow transformers for high-resolution image synthesis},
  author={Esser, Patrick and Kulal, Sumith and Blattmann, Andreas and Entezari, Rahim and M{\"u}ller, Jonas and Saini, Harry and Levi, Yam and Lorenz, Dominik and Sauer, Axel and Boesel, Frederic and others},
  booktitle={Forty-first international conference on machine learning},
  year={2024}
}

@article{xu_imagereward_2023,
  title={Imagereward: Learning and evaluating human preferences for text-to-image generation},
  author={Xu, Jiazheng and Liu, Xiao and Wu, Yuchen and Tong, Yuxuan and Li, Qinkai and Ding, Ming and Tang, Jie and Dong, Yuxiao},
  journal={Advances in Neural Information Processing Systems},
  volume={36},
  pages={15903--15935},
  year={2023}
}

@inproceedings{rombach_highresolution_2022,
  title={High-resolution image synthesis with latent diffusion models},
  author={Rombach, Robin and Blattmann, Andreas and Lorenz, Dominik and Esser, Patrick and Ommer, Bj{\"o}rn},
  booktitle={Proceedings of the IEEE/CVF conference on computer vision and pattern recognition},
  pages={10684--10695},
  year={2022}
}

@inproceedings{podell_sdxl_2023,
  author       = {Dustin Podell and
                  Zion English and
                  Kyle Lacey and
                  Andreas Blattmann and
                  Tim Dockhorn and
                  Jonas M{\"{u}}ller and
                  Joe Penna and
                  Robin Rombach},
  title        = {{SDXL:} Improving Latent Diffusion Models for High-Resolution Image
                  Synthesis},
  booktitle    = {The Twelfth International Conference on Learning Representations,
                  {ICLR} 2024, Vienna, Austria, May 7-11, 2024},
  publisher    = {OpenReview.net},
  year         = {2024},
  url          = {https://openreview.net/forum?id=di52zR8xgf},
  bibsource    = {dblp computer science bibliography, https://dblp.org}
}

@article{ghosh_geneval_2023,
  title={Geneval: An object-focused framework for evaluating text-to-image alignment},
  author={Ghosh, Dhruba and Hajishirzi, Hannaneh and Schmidt, Ludwig},
  journal={Advances in Neural Information Processing Systems},
  volume={36},
  pages={52132--52152},
  year={2023}
}

@article{wu_human_2023,
  author       = {Xiaoshi Wu and
                  Yiming Hao and
                  Keqiang Sun and
                  Yixiong Chen and
                  Feng Zhu and
                  Rui Zhao and
                  Hongsheng Li},
  title        = {Human Preference Score v2: {A} Solid Benchmark for Evaluating Human
                  Preferences of Text-to-Image Synthesis},
  journal      = {CoRR},
  volume       = {abs/2306.09341},
  year         = {2023},
  url          = {https://doi.org/10.48550/arXiv.2306.09341},
  doi          = {10.48550/ARXIV.2306.09341},
  eprinttype   = {arXiv},
  eprint       = {2306.09341},
  bibsource    = {dblp computer science bibliography, https://dblp.org}
}

@article{skalse_defining_2025,
  title={Defining and characterizing reward gaming},
  author={Skalse, Joar and Howe, Nikolaus and Krasheninnikov, Dmitrii and Krueger, David},
  journal={Advances in Neural Information Processing Systems},
  volume={35},
  pages={9460--9471},
  year={2022}
}

@article{kamath_geneval_2025,
  author       = {Amita Kamath and
                  Kai{-}Wei Chang and
                  Ranjay Krishna and
                  Luke Zettlemoyer and
                  Yushi Hu and
                  Marjan Ghazvininejad},
  title        = {GenEval 2: Addressing Benchmark Drift in Text-to-Image Evaluation},
  journal      = {CoRR},
  volume       = {abs/2512.16853},
  year         = {2025},
  url          = {https://doi.org/10.48550/arXiv.2512.16853},
  doi          = {10.48550/ARXIV.2512.16853},
  eprinttype   = {arXiv},
  eprint       = {2512.16853},
  bibsource    = {dblp computer science bibliography, https://dblp.org}
}

@article{kirstain_pickapic_2023,
  title={Pick-a-pic: An open dataset of user preferences for text-to-image generation},
  author={Kirstain, Yuval and Polyak, Adam and Singer, Uriel and Matiana, Shahbuland and Penna, Joe and Levy, Omer},
  journal={Advances in neural information processing systems},
  volume={36},
  pages={36652--36663},
  year={2023}
}

@inproceedings{betker_improving_,
	title = {Improving image generation with better captions},
	author = {Betker, James and Goh, Gabriel and Jing, Li and Brooks, Tim and Wang, Jianfeng and Li, Linjie and Ouyang, Long and Zhuang, Juntang and Lee, Joyce and Guo, Yufei and Manassra, Wesam and Dhariwal, Prafulla and Chu, Casey and Jiao, Yunxin and Ramesh, Aditya},
	year = {2023},
	url = {https://cdn.openai.com/papers/dall-e-3.pdf},
	note = {Accessed: July 22, 2026},
}

@article{deng_emerging_2025,
  author       = {Chaorui Deng and
                  Deyao Zhu and
                  Kunchang Li and
                  Chenhui Gou and
                  Feng Li and
                  Zeyu Wang and
                  Shu Zhong and
                  Weihao Yu and
                  Xiaonan Nie and
                  Ziang Song and
                  Shi Guang and
                  Haoqi Fan},
  title        = {Emerging Properties in Unified Multimodal Pretraining},
  journal      = {CoRR},
  volume       = {abs/2505.14683},
  year         = {2025},
  url          = {https://doi.org/10.48550/arXiv.2505.14683},
  doi          = {10.48550/ARXIV.2505.14683},
  eprinttype   = {arXiv},
  eprint       = {2505.14683},
  bibsource    = {dblp computer science bibliography, https://dblp.org}
}

@article{singh_openai_2025,
  author       = {OpenAI},
  title        = {OpenAI {GPT-5} System Card},
  journal      = {CoRR},
  volume       = {abs/2601.03267},
  year         = {2026},
  url          = {https://doi.org/10.48550/arXiv.2601.03267},
  doi          = {10.48550/ARXIV.2601.03267},
  eprinttype   = {arXiv},
  eprint       = {2601.03267},
  bibsource    = {dblp computer science bibliography, https://dblp.org}
}

@inproceedings{zhang_adding_2023,
  title={Adding conditional control to text-to-image diffusion models},
  author={Zhang, Lvmin and Rao, Anyi and Agrawala, Maneesh},
  booktitle={Proceedings of the IEEE/CVF international conference on computer vision},
  pages={3836--3847},
  year={2023}
}

\newpage
\section*{Supplemental Material}

\setcounter{figure}{0}
\setcounter{table}{0}
\renewcommand{\thefigure}{S\arabic{figure}}
\renewcommand{\thetable}{S\arabic{table}}
\setcounter{equation}{0}
\renewcommand{\theequation}{S\arabic{equation}}

\setcounter{section}{0}
\renewcommand{\thesection}{S\arabic{section}}
\setcounter{subsection}{0}
\renewcommand{\thesubsection}{S\arabic{subsection}}
\setcounter{subsubsection}{0}
\renewcommand{\thesubsubsection}{S\arabic{subsubsection}}

\subsection{Analysis of Intermediate Rewards}
\label{sec:analysis}

\begin{figure}[h]
  \centering
  \newcommand{\corrwidth}{0.95\columnwidth}
  \includegraphics[width=\corrwidth]{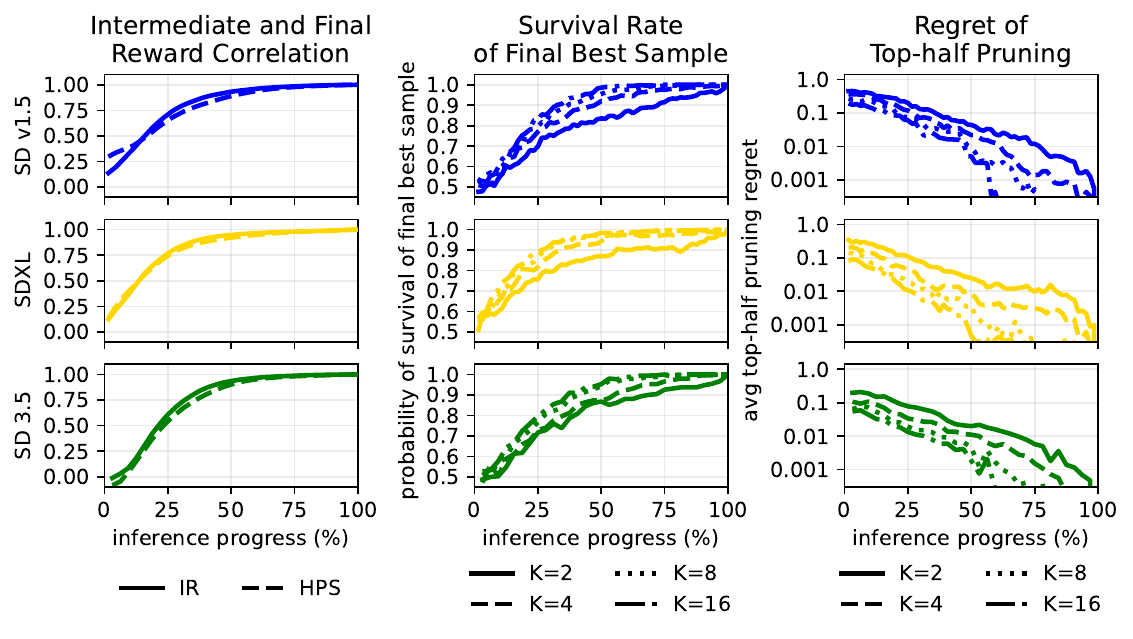}
  \caption{\textbf{How informative are intermediate rewards?}
  {\bf Left}: correlation between intermediate rewards $r(\hat{x}_0(x_t))$ and final rewards $r(x_0)$ over inference progress for three backbones; ImageReward provides higher correlation than \HPS \ across most steps.
  {\bf Middle}: probability that the final best sample (by final reward) survives when pruning from $K$ candidates to $K/2$ at a given progress point.
  {\bf Right}: regret (difference between final best sample and surviving best sample) when pruning from $K$ candidates to $K/2$}
  \label{fig:intermediate_reward_correlation}
\end{figure}

The effectiveness of \PSP \ hinges on whether intermediate reward rankings preserve the best final seeds.
\cref{fig:intermediate_reward_correlation} (left) shows that intermediate rewards become informative relatively early: correlations rise rapidly and remain high across backbones.
Notably, ImageReward exhibits stronger correlation than \HPS \ at most progress points, which helps explain why IR-guided selection works better for \PSP \ and translates more reliably to GenEval improvements in \cref{tab:main_models_samplers}.

Correlation, however, is not the full story: \PSP \ needs intermediate \emph{rankings} to preserve top candidates under pruning.
\cref{fig:intermediate_reward_correlation} (middle) directly measures this by estimating the probability that the final best sample remains in the survivor set after pruning from $K$ to $K/2$ based on intermediate scores at a given step.
For the pruning used in our default schedule, survival probabilities are high across models (80\% survival rate for pruning from 8 at 25\% inference and 90\% for pruning from 4 at 50\% inference), supporting the core assumption behind \PSP. Even if the best seed does not survive, \cref{fig:intermediate_reward_correlation} (right) shows the average regret of the same pruning operations, the difference between the final reward of the true best seed and best surviving seed. Our default schedule has regret close to 0.01 at both our pruning times, except for the weaker SD v1.5 model.

\newpage
\subsection{Generalization of tuned schedules and reward saturation.}
\label{sec:analysis}
\begin{figure}[h]
  \centering
  \newcommand{\distwidth}{0.95\columnwidth}
  \includegraphics[width=\distwidth]{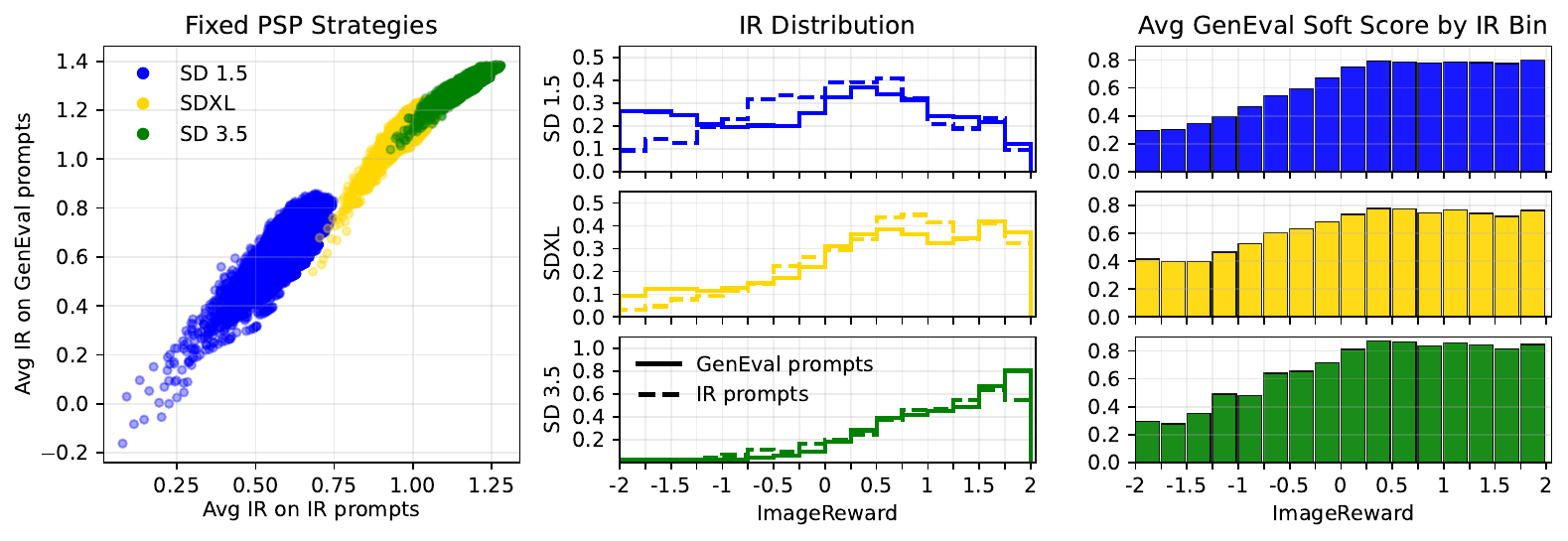}
  \caption{\textbf{Generalization of tuned schedules and reward saturation.}
  Left: reward achieved by each schedule on the tuning set versus on GenEval, showing positive correlation and suggesting limited overfitting.
  Middle: reward distributions for representative schedules remain similar across prompt sets.
  Right: GenEval versus reward (binned by IR), illustrating diminishing returns at high IR where further improvements in the guidance metric translate weakly to GenEval.}
  \label{fig:reward_distributions_and_binned_geneval}
\end{figure}
\newpage

\subsection{Experiments with HPS Guidance}

\begin{table}[h!]
  \caption{\textbf{Results under HPS guidance.}
  Guidance reward (\HPS) and GenEval when \HPS\ is used as the guidance signal,
  under matched effective compute.}
  \label{tab:hps_guidance}
  \centering
  \setlength{\tabcolsep}{4pt}

  \begin{tabular}{@{}llcccc@{}}
    \toprule
    Model & Sampler & $T$ & $\bar{N}$ & HPS $\uparrow$ & GenEval $\uparrow$ \\
    \midrule

    SD v1.5 & Standard                                & 64 & 1 & 0.257 & 0.434 \\
    SD v1.5 & Best-of-N                               & 64 & 4 & 0.282 & 0.525 \\
    SD v1.5 & FK-Steering \cite{singhal_general_2025} & 64 & 4 & 0.280 & 0.511 \\
    SD v1.5 & DSearch \cite{li_dynamic_2025}          & 64 & 4 & \textbf{0.293} & 0.505 \\
    \midrule
    SD v1.5 & \PSP                                    & 64 & 4 & 0.288 & \textbf{0.541} \\
    \midrule\midrule

    SDXL & Standard                                   & 64 & 1 & 0.275 & 0.529 \\
    SDXL & Best-of-N                                  & 64 & 4 & 0.298 & 0.617 \\
    SDXL & FK-Steering \cite{singhal_general_2025}    & 64 & 4 & 0.299 & 0.590 \\
    SDXL & DSearch \cite{li_dynamic_2025}             & 64 & 4 & \textbf{0.311} & 0.580 \\
    \midrule
    SDXL & \PSP                                       & 64 & 4 & 0.304 & \textbf{0.625} \\
    \midrule\midrule

    SD 3.5 & Standard                                 & 32 & 1 & 0.297 & 0.713 \\
    SD 3.5 & Best-of-N                                & 32 & 4 & 0.312 & 0.748 \\
    SD 3.5 & FK-Steering \cite{singhal_general_2025}  & 32 & 4 & 0.302 & 0.729 \\
    \midrule
    SD 3.5 & \PSP                                     & 32 & 4 & \textbf{0.316} & \textbf{0.753} \\
    \bottomrule
  \end{tabular}
\end{table}

\newpage
\subsection{Stochasticity for Stable Diffusion 3.5}
\label{sec:stochastic-sd35}

\begin{figure}[h]
    \centering
    \includegraphics[width=0.95\linewidth]{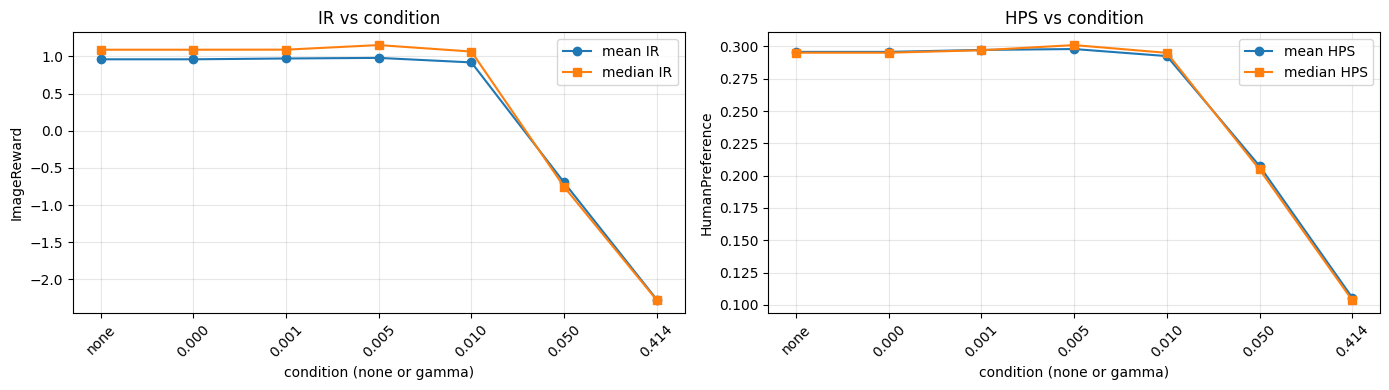}
    \caption{
    SD3.5 stochasticity calibration on the held-out IR benchmark.
    We compare deterministic (\texttt{none}), wrapper sanity check (\(\gamma=0\)),
    and increasing controlled noise levels (\(\gamma=0.001, 0.005, 0.01, 0.05, 0.414\)).
    The selected operating point is \(\gamma=0.005\), the highest tested stochasticity
    that does not reduce final reward.
    }
    \label{fig:sd35-gamma-sweep}
\end{figure}

We start from the SD3.5 flow-matching Euler solver. Let $x_i$ be the latent at step $i$,
$\sigma_i$ the corresponding noise level, and $v_\theta(x_i,\sigma_i,c)$ the model prediction.
With deterministic sampling, the update is
\begin{equation}
x_{i+1}
=
x_i + (\sigma_{i+1}-\sigma_i)\,v_\theta(x_i,\sigma_i,c).
\label{eq:sd35_det}
\end{equation}

\paragraph{Standard \texttt{stochastic\_sampling} implementation.}
In the default SD3.5 scheduler implementation, enabling \texttt{stochastic\_sampling}
switches the step rule to
\begin{align}
\hat{x}_0 &= x_i - \sigma_i\,v_\theta(x_i,\sigma_i,c),\\
x_{i+1} &= (1-\sigma_{i+1})\,\hat{x}_0 + \sigma_{i+1}\,\epsilon_i,\quad
\epsilon_i \sim \mathcal{N}(0,I).
\label{eq:sd35_builtin_stoch}
\end{align}
This is a binary on/off mode: the noise magnitude is implicitly tied to
$\sigma_{i+1}$ by the solver itself, with no direct user control over strength
(or schedule) beyond toggling the flag. In our experiments, this lack of control
consistently produced worse downstream metrics, so we do not use this built-in mode.

\paragraph{Our controlled stochastic modification.}
Instead, we keep the deterministic SD3.5 Euler step in Eq.~\eqref{eq:sd35_det} and
inject noise \emph{after} the base update:
\begin{align}
x_{i+1}^{\mathrm{det}}
&=
x_i + (\sigma_{i+1}-\sigma_i)\,v_\theta(x_i,\sigma_i,c),\\
x_{i+1}
&=
x_{i+1}^{\mathrm{det}} + \sigma_{\mathrm{noise},i}\,\epsilon_i,\quad
\epsilon_i \sim \mathcal{N}(0,I),
\end{align}
with
\begin{equation}
\sigma_{\mathrm{noise},i}
=
\sigma_i\sqrt{(1+\gamma)^2-1}\,s_{\mathrm{noise}}.
\label{eq:sd35_controlled_noise}
\end{equation}
We set $s_{\mathrm{noise}}=1$, $s_{t,\min}=0$, and $s_{t,\max}=+\infty$, and tune
$\gamma$ directly. Operationally, $\gamma$ is clamped to
\begin{equation}
\gamma \in [0,\sqrt{2}-1].
\end{equation}
Hence, unlike Eq.~\eqref{eq:sd35_builtin_stoch}, our method provides explicit and
continuous control of stochasticity: $\gamma=0$ recovers deterministic SD3.5 exactly,
while increasing $\gamma$ increases exploration in a predictable way.

\paragraph{Gamma sweep on held-out tuning data.}
Figure~\ref{fig:sd35-gamma-sweep} summarizes our sweep over
\(
\gamma \in \{\texttt{none}, 0, 0.001, 0.005, 0.01, 0.05, 0.414\}
\)
on the held-out IR benchmark (used only for hyperparameter tuning).
Here, \texttt{none} means the fully deterministic SD3.5 code path (no stochasticity),
while \(\gamma=0\) is a sanity check for our stochastic wrapper with zero injected noise.
Both sanity checks should be consistent up to numerical noise. As shown in the sweep,
\(\gamma=0.005\) is the largest noise level that does not degrade final image-reward
performance, so we use it as our stochastic setting.

\newpage
\subsection{\PSP \ Schedule Grid Search}
\label{sec:gridsearch}

\begin{table}[h]
  \caption{Effect of tuning on \PSP, separated by algorithm (Scheduled vs Dynamic). Tuned columns use IR-tuned variants for IR/GenEval (IR) and HPS-tuned variants for HPS/GenEval (HPS).}
  \label{tab:tuning_scheduled_dynamic}
  \centering
  \setlength{\tabcolsep}{4pt}

  \begin{tabular}{@{}lcccccccc@{}}
    \toprule
    \multirow{2}{*}{Model} &
    \multicolumn{2}{c}{IR} &
    \multicolumn{2}{c}{HPS} &
    \multicolumn{2}{c}{GenEval (IR)} &
    \multicolumn{2}{c}{GenEval (HPS)} \\
    \cmidrule(lr){2-3}\cmidrule(lr){4-5}\cmidrule(lr){6-7}\cmidrule(lr){8-9}
     & \PSP & Tuned & \PSP & Tuned & \PSP & Tuned & \PSP & Tuned \\
    \midrule

    SD v1.5 & 0.827 & 0.836 & 0.288 & 0.288 & 0.574 & 0.571 & 0.541 & 0.541 \\
    SDXL    & 1.224 & 1.205 & 0.304 & 0.305 & 0.645 & 0.641 & 0.625 & 0.646 \\
    SD3.5   & 1.380 & 1.380 & 0.316 & 0.316 & 0.747 & 0.753 & 0.753 & 0.751 \\
    \bottomrule
  \end{tabular}
\end{table}

We tune the \emph{standard} Progressive Seed Pruning (\PSP) by searching over pruning
schedules on a held-out tuning split (IR benchmark). For each model (SD1.5, SDXL, SD3.5),
we run two separate searches: one guided by ImageReward and one guided by HumanPreference.
The selected schedule is then used as the \PSP \ tuned configuration for that model/metric setting in \cref{sec:tuning}.

\paragraph{Compute-efficient search protocol.}
To make this search tractable, we separate generation from schedule evaluation. For each prompt in IR Benchmark, we run 32 seeds and store per-step metrics
(e.g., \texttt{prompt\_id}, \texttt{seed}, \texttt{step}, \texttt{image\_reward},
\texttt{human\_preference}) in a CSV file. Then, grid search evaluates \PSP \
candidates by replaying pruning decisions on these cached trajectories, without running
the diffusion model again. This lets us evaluate a large set of candidate schedules
under fixed samples/trajectories, while constraining the search to at most four pruning events.

\paragraph{What is searched.}
A \PSP \ schedule is parameterized by:
\begin{itemize}
    \item initial particle count \(k_{\mathrm{init}}\),
    \item pruning times \((t_1,\dots,t_m)\) with \(1 \le m \le 4\),
    \item survivor counts \((k_1,\dots,k_m)\), where \(k_j\) is the number of particles kept after pruning at \(t_j\).
\end{itemize}
The search enforces:
\begin{itemize}
    \item strictly increasing pruning times \(t_1 < \cdots < t_m\),
    \item strictly decreasing particle counts \(k_{\mathrm{init}} > k_1 > \cdots > k_m\),
    \item total compute budget no larger than Best-of-\(N\) baseline budget (\(N=4\)).
\end{itemize}

\paragraph{Budget constraint.}
Let total denoising length be \(T\). \\
For schedule
\(\big(k_{\mathrm{init}}, (t_1,\dots,t_m), (k_1,\dots,k_m)\big)\), the particle-step cost is:
\begin{equation}
C
=
k_{\mathrm{init}}\, t_1
+
\sum_{j=1}^{m-1} k_j (t_{j+1}-t_j)
+
k_m (T-t_m).
\end{equation}
We keep only schedules with
\begin{equation}
C \le 4T.
\end{equation}
Thus, all \PSP \ candidates are compared under the same compute envelope as Best-of-4.

\paragraph{Search grids used in our scripts.}
Across all six benchmark-IR tuning runs \\(SD1.5/SDXL/SD3.5 \(\times\) IR/HPS guidance), we use:
\begin{itemize}
    \item \textbf{Guidance metric:} either \texttt{image\_reward} or \texttt{human\_preference}.
    \item \textbf{Logical seeds:} \(\{0,1\}\) (reported metric is averaged across these two seed windows).
    \item \textbf{Max pruning events:} \(m \le 4\).
    \item \textbf{Candidate survivor counts:} \(\{16,12,8,4,3,2,1\}\) (also provides candidates for \(k_{\mathrm{init}}\)).
    \item \textbf{Candidate pruning times:}
    \begin{itemize}
        \item SD1.5 / SDXL (\(T=64\)): \(\{4,8,12,16,20,24,28,32,36,40,44,48,52,56,60\}\),
        \item SD3.5 (\(T=32\)): \(\{2,4,6,8,10,12,14,16,18,20,22,24,26,28,30\}\).
    \end{itemize}
\end{itemize}

\paragraph{Selection criterion.}
For each valid schedule, we simulate \PSP \ selection per prompt and keep the final sample
at step \(T\). Schedules are ranked by mean final metric over prompts and averaged over the two logical seeds.
The top-ranked schedule is the tuned \PSP \ schedule for that model/reward setting, tuned on the IR benchmark dataset. 

Precomputing all intermediate rewards (IR and HPS) for SD 3.5 takes 39 H200 hours for the 100 prompts in the IR benchmark and 216 H200 hours (9 days) for the 553 prompts in the Geneval benchmark. THe gridsearch on the cached values takes 40min, without need for any GPU.

\newpage
\subsection{Prompt Rewriting}
\label{sec:rewriting}

Prompt rewriting is used to increase semantic diversity \emph{without} changing the underlying GenEval constraints.
The rewriting instruction (shown below) was issued to \textbf{ChatGPT 5.2 Extended Thinking} on \textbf{Feb 28, 2026}, and it took it \textbf{6min 42s} of thinking to generate it. The resulting rewritten prompt set is used as a fixed preprocessing artifact.

\paragraph{How rewriting is used in search.}
For each original GenEval prompt \(p\), we generate 24 constraint-preserving rewrites
\(\{p^{(0)},\dots,p^{(23)}\}\). During search/evaluation, we keep the \emph{same initial noise seed}
for all rewrites of the same base prompt. This controls stochastic variation from noise and makes comparisons across rewrites primarily reflect
prompt wording/composition effects rather than different random initializations.

\paragraph{Why GenEval scoring is still valid.}
GenEval metrics are object-centric (required objects/counts/attributes/relations).
Because rewrites are constrained to preserve exactly those semantic constraints, generated images remain
compatible with the same GenEval checks. In other words, rewriting changes phrasing/style/context,
but not the measurable task requirements (object presence, counts, colors, and spatial relations).

\paragraph{Prompt:}

\begin{verbatim}
You are generating prompt expansions for GenEval while preserving 
the original scoring constraints.

Goal:
- For each original GenEval prompt entry, create exactly 24 prompt 
expansions.
- Expansions should be semantically equivalent and 
constraint-preserving, but diverse enough to encourage different 
images under the same initial noise.

Input file:
- geneval_metadata.jsonl (one JSON object per line).

Output file:
- geneval_metadata_multiprompts.json
- Must be a valid JSON array.

Required output schema (per original prompt entry):
{
  "prompt_id": <int>,
  "tag": <string>,
  "include": <original include array>,
  "exclude": <original exclude array, if present>,
  "original_prompt": <string>,
  "expansions": [
    {"prompt_expansion_id": 0, "prompt": <string>},
    {"prompt_expansion_id": 1, "prompt": <string>},
    ...
    {"prompt_expansion_id": 23, "prompt": <string>}
  ]
}

Hard requirements:
1) Keep constraints identical to original entry:
   - same object classes
   - same counts
   - same color requirements
   - same positional relations (left/right/above/below) when present
   - never add contradictory constraints
2) Create exactly 24 expansions for each prompt_id.
3) Use unique prompt_expansion_id values 0..23.
4) Make expansions meaningfully different from each other:
   - vary composition, camera framing, distance, angle, lighting, 
   scene context, background style
   - do NOT change required semantic constraints
5) Prompt length:
   - target 35-45 words
   - hard cap 50 words per expanded prompt
6) Output strictly valid JSON only (no markdown, no commentary).

Diversity guidance:
- Mix close-up, medium, wide shots.
- Vary environment (studio, outdoor, indoor, urban, natural) where 
compatible.
- Vary lighting (soft daylight, overcast, warm indoor, dramatic 
side light), without changing semantics.
- Vary descriptive style while preserving all required
objects/attributes/relations.

Validation checklist before final output:
- JSON parses.
- Number of entries equals number of input prompts.
- Each entry has 24 expansions.
- Expansion IDs are exactly 0..23.
- All prompts are <= 50 words.
- Constraints preserved for each prompt_id.
\end{verbatim}

\newpage
\subsection{Experimental Details}
\label{sec:sup_exps}

All results for deterministic strategies (regular inference, \BoN \  and \PSP) are computed from pre-generated reward trajectories for efficiency and reproducibility.
For each model and prompt, we cache per-step rewards across multiple initial noise seeds. We then use a logical seed to determine the initial noise seeds of a run and evaluate the results that would be achieved by regular inference, \BoN, and \PSP. For \PSP, evaluation is done by replaying pruning/selection decisions on those cached trajectories.
This avoids rerunning diffusion sampling for every strategy variant and makes broad schedule sweeps tractable, as discussed in \cref{sec:gridsearch}.

\subsubsection{Baselines and Main Comparison}
\label{sec:sup_exps_main}

\paragraph{Models and prompts.}
We evaluate Stable Diffusion v1.5, SDXL, and Stable Diffusion 3.5 on GenEval prompts.
Results are reported for ImageReward (IR) guidance in \cref{tab:main_models_samplers} and HumanPreference (HPS) guidance in \cref{tab:hps_guidance}.

\paragraph{Methods compared.}
We compare:
\begin{itemize}
    \item \textbf{Standard}: single-sample generation (\(N=1\)).
    \item \textbf{Best-of-\(N\)}: \(N=4\), selecting the final sample with the best guidance score.
    \item \textbf{FK-Steering}: sequential Monte Carlo with \(K=4\) particles and \(\lambda=10\), using \texttt{potential\_type=max} and non-adaptive resampling.  
    For SD v1.5/SDXL (\(T=64\)): resampling every 12 steps over \(t\in[12,48]\), with \(\eta=1.0\).  
    For SD 3.5 (\(T=32\)): resampling every 6 steps over \(t\in[6,24]\), with stochastic step-wrapping \(\gamma=0.005\).
    \item \textbf{BFS}: BFS is FK-Steering with only the resampling method and tempering schedule altered; the remainder of the sequential Monte Carlo procedure is unchanged. We retain \(K=4\) particles, \(\lambda=10\), \texttt{potential\_type=max}, and the same resample windows (\(t\in[12,48]\) every 12 steps for \(T=64\); \(t\in[6,24]\) every 6 steps for \(T=32\)), and modify two settings. First, resampling: FK-Steering uses multinomial resampling, which draws each surviving particle independently in proportion to its weight and therefore injects higher sampling variance (particles can be duplicated or dropped erratically); BFS uses SSP resampling (\texttt{resampling=ssp}, Srinivasan Sampling Process), a low-variance offspring-allocation scheme in which offspring counts more closely track particle weights, improving population diversity under finite \(K\). Second, the tempering schedule, which controls how sharply weights are annealed over the trajectory: FK-Steering keeps it constant (\texttt{tempering\_schedule=constant}), applying the same selection pressure at every resample regardless of timestep; BFS uses an increasing schedule instead (\texttt{tempering\_schedule=increase}), so selection is softer early (when \(\hat{x}_0\) estimates are less reliable) and sharper later (when they are more informative). SD 3.5 (\(T=32\)) uses stochastic step-wrapping \(\gamma=0.005\).
    \item \textbf{DSearch}: for SD v1.5/SDXL (\(T=64\)) and SD 3.5 (\(T=32\)), guidance in \{IR, HPS\}, and search hyperparameters:\\
    \(\texttt{num\_images}=1\),
    \(\texttt{bs}=1\),
    \(\texttt{duplicate\_size}=1\),
    \(\texttt{w}=2\),
    \(\texttt{oversamplerate}=2\),
    \(\texttt{search\_schudule}=\texttt{all}\),
    \(\texttt{drop\_schudule}=\texttt{exponential}\),\\
    \(\texttt{replacerate}=0\),
    \(\texttt{variant}=\texttt{PM}\),
    and \(\eta=1.0\). SD 3.5 additionally uses stochastic step-wrapping \(\gamma=0.005\).
    \item \textbf{Noise Trajectory Search (NTS)}: local \(\epsilon\)-greedy noise-trajectory search guided by ImageReward, with \(K=2\) refinement rounds and \(N=2\) candidates per round at each of the \(T-1\) transitions. We use \(\epsilon=0.4\) and \(\lambda=0.15\). For SD v1.5/SDXL: \(T=64\), guidance scale \(7.5\), \(\eta=1.0\). For SD 3.5: \(T=32\), guidance scale \(7.0\), stochastic step-wrapping \(\gamma=0.005\).
    \item \textbf{Rollover Budget Forcing (RBF)}: particle-based sampling with a rolling per-step NFE budget capped at \(\texttt{max\_nfe}\), \(\texttt{batch\_size}=2\), IR reward.
    SD v1.5/SDXL: \(\texttt{init\_n\_particles}=8\), \(\texttt{max\_nfe}=256\), \(T=64\), guidance scale \(7.5\), stochastic DDIM with \(\eta=1.0\) on the native VP schedule.
    SD 3.5: \(\texttt{init\_n\_particles}=4\), \(\texttt{max\_nfe}=128\), \(T=32\), guidance scale \(7.0\), SDE integration (\(\texttt{diffusion\_coefficient}=\texttt{square}\), \(\texttt{diffusion\_norm}=3.0\)) on a VP-converted flow-matching schedule.
    \item \textbf{SVDD}: we run SVDD-PM with \(\texttt{duplicate\_size}=4\) (4 candidates/particles per prompt) and ImageReward guidance. 
    For SD v1.5 (\(T=64\)): guidance scale \(7.5\), resolution \(512\times512\), \(\eta=1.0\), \(\texttt{variant}=\texttt{PM}\).
    For SDXL (\(T=64\)): guidance scale \(7.5\), resolution \(1024\times1024\), with true duplicate-guided SVDD resampling configured as
    \(\texttt{num\_particles}=4\), \(\lambda=10\), \(\texttt{potential\_type}=\texttt{diff}\), \(\texttt{resampling}=\texttt{multinomial}\), \(\texttt{resample\_frequency}=1\), non-adaptive resampling, \(\texttt{resampling\_t\_start}=0\), \(\texttt{resampling\_t\_end}=64\), and lastly \(\texttt{tempering\_schedule}=\texttt{constant}\).
    For SD 3.5 (\(T=32\)): guidance scale \(7.0\), resolution \(1024\times1024\), the same SVDD resampling configuration as before (\(\texttt{resampling\_t\_end}=32\)), plus stochastic step-wrapping with \(\gamma=0.005\).
    \item \textbf{\PSP \ (ours)}: default schedules are
    \begin{itemize}
        \item SD v1.5/SDXL: \(k_{\mathrm{init}}=8,\; t=(16,32),\; k=(4,2),\; T=64\);
        \item SD 3.5: \(k_{\mathrm{init}}=8,\; t=(8,16),\; k=(4,2),\; T=32\).
    \end{itemize}
\end{itemize}

\paragraph{Evaluation protocol.}
Each method is evaluated on GenEval by selecting one final sample per prompt (the one with highest reward guidance), then aggregating:
\begin{itemize}
    \item mean final reward guidance (IR or HPS)
    \item GenEval overall score (task-wise average correctness).
\end{itemize}
For all results in \cref{tab:main_models_samplers} and \cref{tab:hps_guidance}, we report average results across 3 different runs. For standard/BoN/\PSP, we precompute 32 initial noise seeds per prompt and use different non-overlapping groups of initial noise seeds to report the average results across different runs.

\subsubsection{Scaling with Compute}
\label{sec:sup_exps_scaling}

\paragraph{Goal and setup.}
We compare BoN and \PSP \ under larger compute budgets with effective
\(\bar N \in \{2,4,8,16\}\) (optionally including BoN at \(N=1\) as a reference point).
Results are shown for SD v1.5, SDXL, and SD 3.5.

\paragraph{Scaling strategy.}
Starting from a base \PSP \ schedule, we scale \(k_{\mathrm{init}}\) and survivor counts
proportionally with \(\bar N\), enforce proportional pruning constraints, and evaluate each scaled schedule
by replaying on cached trajectories for \BoN \ and \PSP.

\paragraph{Plotted quantities.}
We report:
\begin{itemize}
    \item score vs effective \(\bar N\) (IR or GenEval),
    \item regret vs \(\bar N\), defined as
    \[
      \mathrm{Regret}(\bar N) = \mathrm{BoN}(2\bar N) - \mathrm{\PSP}(\bar N),
    \]
    \item score vs total generation FLOPs.
\end{itemize}

\paragraph{FLOPs accounting.}
Per-step FLOPs are measured from one denoiser forward pass using profiler-based FLOPs accounting
(UNet for SD v1.5/SDXL, diffusion transformer for SD 3.5). Total generation FLOPs are then:
\[
\mathrm{FLOPs}_{\mathrm{total}} = \bar N \times \mathrm{FLOPs}_{\mathrm{step}} \times T,
\]
with \(T=64\) for SD v1.5/SDXL and \(T=32\) for SD 3.5.

\subsubsection{Analysis of Intermediate Rewards}
\label{sec:sup_exps_intermediate}

We analyze how informative intermediate rewards are about final outcomes, using per-step cached trajectories.

\paragraph{Intermediate--final reward correlation.}
We use 32 seeds per prompt. At each denoising step \(t\), we pool all \((\text{prompt},\text{seed})\) pairs and compute Pearson correlation between intermediate and final rewards for the same seed:
\[
\mathrm{corr}\!\left(r_t^{\mathrm{IR}}(p,s),\, r_T^{\mathrm{IR}}(p,s)\right),
\qquad
\mathrm{corr}\!\left(r_t^{\mathrm{HPS}}(p,s),\, r_T^{\mathrm{HPS}}(p,s)\right).
\]
Thus, the curve shows how predictive step-\(t\) reward is of that seed's final reward at \(T\), aggregated over all prompts and all 32 seeds.

\paragraph{Survival rate of the final-best sample under pruning.}
For each \(K \in \{2,4,8,16\}\) and prompt $p$, we form a fixed seed bag of $K$ seeds \(\mathcal{S}_K(p)=\{0,\dots,K-1\}\). Then, let
\[
s_{K,p}^\star = \arg\max_{s \in \mathcal{S}_K(p)} r_T^{\mathrm{IR}}(p,s)
\]
be the final best seed at step \(T\) from the original K seeds in \(\mathcal{S}_K(p)\). At step \(t\), we keep the top half of \(\mathcal{S}_K(p)\) by intermediate IR and check whether \(s_p^\star\) survives. The reported curve is the frequency of this survival over all prompts per timestep.

\paragraph{Regret after pruning.}
Using the same fixed bag \(\mathcal{S}_K(p)\), top-half survivors are selected at step \(t\) by intermediate IR, and regret is computed in final IR space:
\[
\mathrm{Regret}_K(t)
=
\max_{s \in \mathcal{S}_K(p)} r_T^{\mathrm{IR}}(p,s)
-
\max_{s \in \mathrm{TopHalf}_t(\mathcal{S}_K(p))} r_T^{\mathrm{IR}}(p,s).
\]
The plotted value is the average of this quantity across prompts.

\subsubsection{Tuned \PSP \ Schedules}
\label{sec:sup_exps_tuned_pps}

The tuned \PSP \ schedules are selected by the grid-search procedure described in \cref{sec:gridsearch}.
We report tuned results using two logical seeds (that define the group of initial noise seeds) so that, with 32 precomputed initial-noise seeds per prompt,
we can always form two non-overlapping windows even for the largest initial pool (\(k_{\mathrm{init}}=16\)). We use prompts from Benchmark IR for the search and chose the best pruning schedule per model/reward pair.

\paragraph{Chosen schedules (tuned setting).}
\begin{itemize}
    \item \textbf{SD v1.5 (IR):}
    \(k_{\mathrm{init}}=8,\; t=(20,36,56),\; k=(3,2,1),\; T=64\).
    \item \textbf{SD v1.5 (HPS):}
    \(k_{\mathrm{init}}=16,\; t=(4,8,20,28),\; k=(12,4,3,2),\; T=64\).
    \item \textbf{SDXL (IR):}
    \(k_{\mathrm{init}}=16,\; t=(4,8,24,40),\; k=(8,4,3,2),\; T=64\).
    \item \textbf{SDXL (HPS):}
    \(k_{\mathrm{init}}=12,\; t=(8,12,24),\; k=(8,4,2),\; T=64\).
    \item \textbf{SD 3.5 (IR):}
    \(k_{\mathrm{init}}=8,\; t=(8,12,22,30),\; k=(4,3,2,1),\; T=32\).
    \item \textbf{SD 3.5 (HPS):}
    \(k_{\mathrm{init}}=8,\; t=(8,12,22,30),\; k=(4,3,2,1),\; T=32\).
\end{itemize}

\subsubsection{Comparison to Finetuned Models}
\label{sec:sup_exps_dpo}

\paragraph{Models.}
We include two DPO-finetuned generators:
\begin{itemize}
    \item a DPO-finetuned SD v1.5 model,
    \item a DPO-finetuned SDXL model.
\end{itemize}

\paragraph{Protocol.}
For each finetuned model, we run the same offline selection pipeline from precomputed inference trajectories as in the base-model setting,
comparing standard inference, BoN (\(N=4\)), and \PSP \ under matched logical-seed windows.

\paragraph{Guidance and schedule.}
In these finetuned-model comparisons, guidance is IR-based.
\PSP \ is evaluated with \(k_{\mathrm{init}}=8\), cutoff times \((16,32)\), survivors \((4,2)\), and \(T=64\). Results are averaged over 3 logical seeds (to generate 3 non-overlapping sets of initial noise seeds).

\newpage

\subsection{Predicted Clean Images ($\hat{x}_0$) Examples}
\label{sec:sup_x0_examples}

To make concrete what the off-the-shelf reward function actually scores at intermediate steps, we visualize the predicted clean images $\hat{x}_0$ that our method feeds to the reward model at different timesteps and across backbones. At each step the reward is not computed on the noisy latent $x_t$ (top row) but on the model's clean-image estimate $\hat{x}_0$ (bottom row), decoded to pixel space. The ImageReward of each estimate is reported below it. Notice how the coarse features of an image, which are crucial for prompt alignment, are defined very early in the denoising process, so even the early $\hat{x}_0$ estimates already expose the layout and object composition that the reward function can act on.

\begin{figure*}[h]
    \centering
    \includegraphics[width=\textwidth]{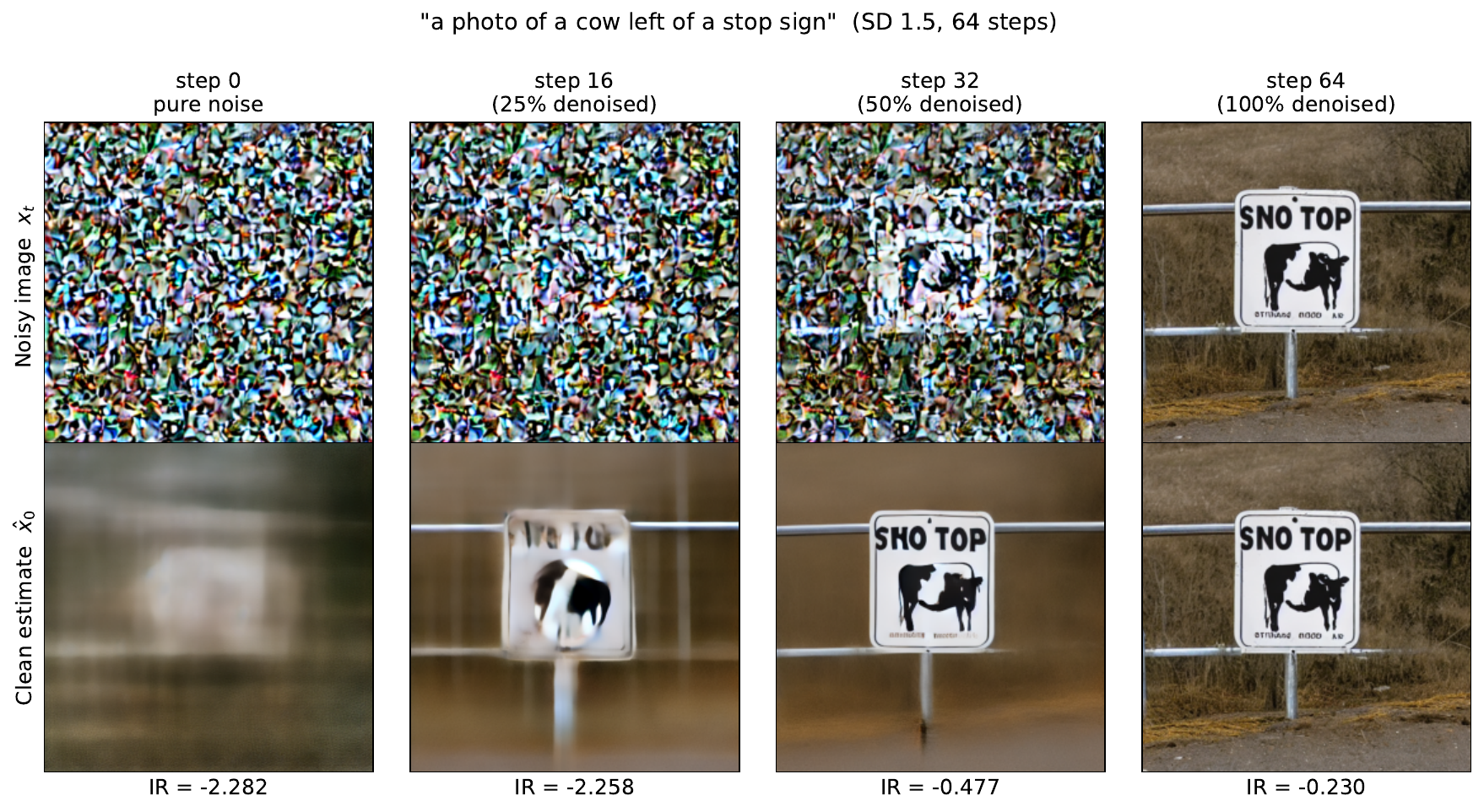}
    \caption{Predicted clean images $\hat{x}_0$ for Stable Diffusion v1.5 (64 steps).}
    \label{fig:sup_x0_sd15}
\end{figure*}

\begin{figure*}[h]
    \centering
    \includegraphics[width=\textwidth]{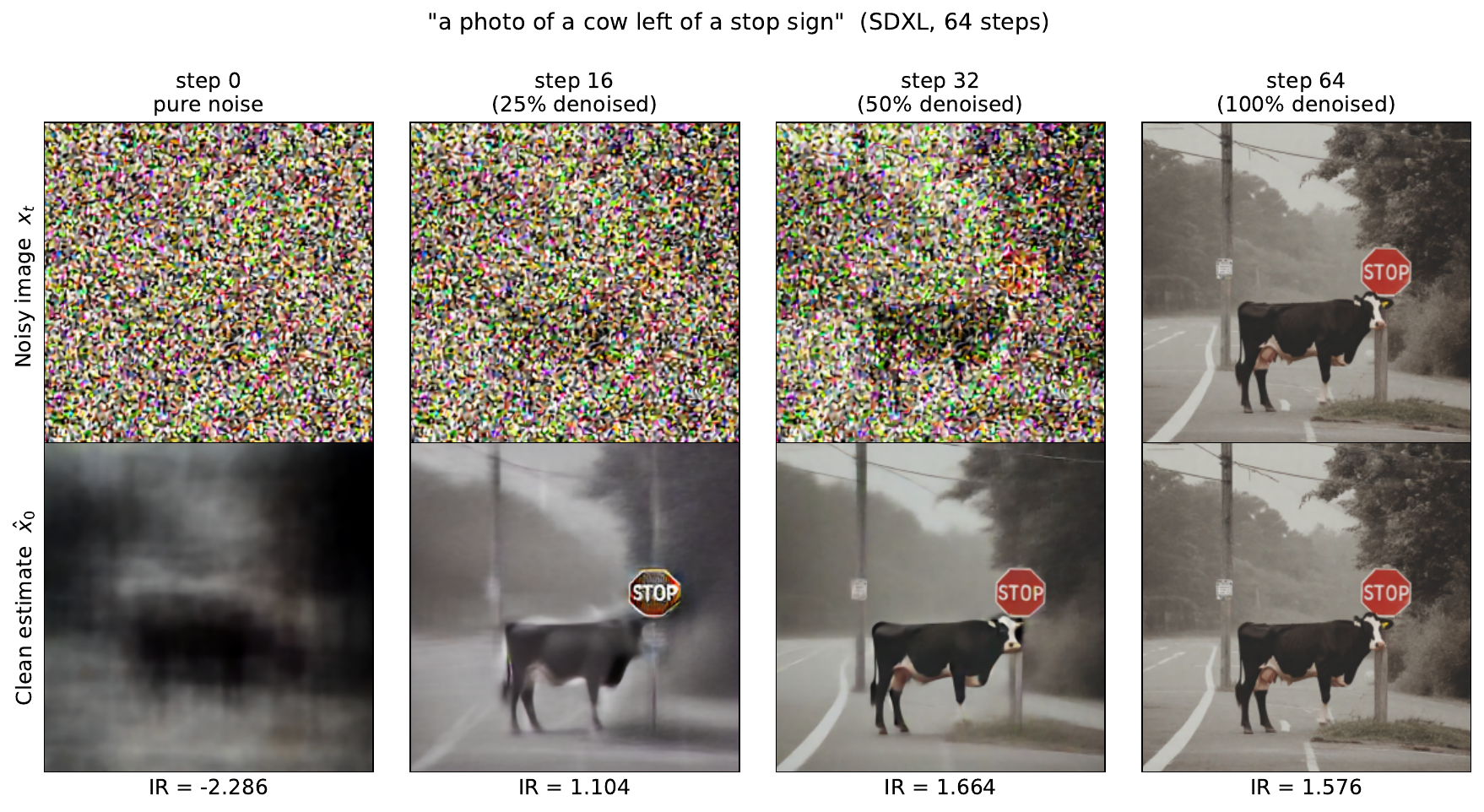}
    \caption{Predicted clean images $\hat{x}_0$ for Stable Diffusion XL (64 steps).}
    \label{fig:sup_x0_sdxl}
\end{figure*}

\begin{figure*}[h]
    \centering
    \includegraphics[width=\textwidth]{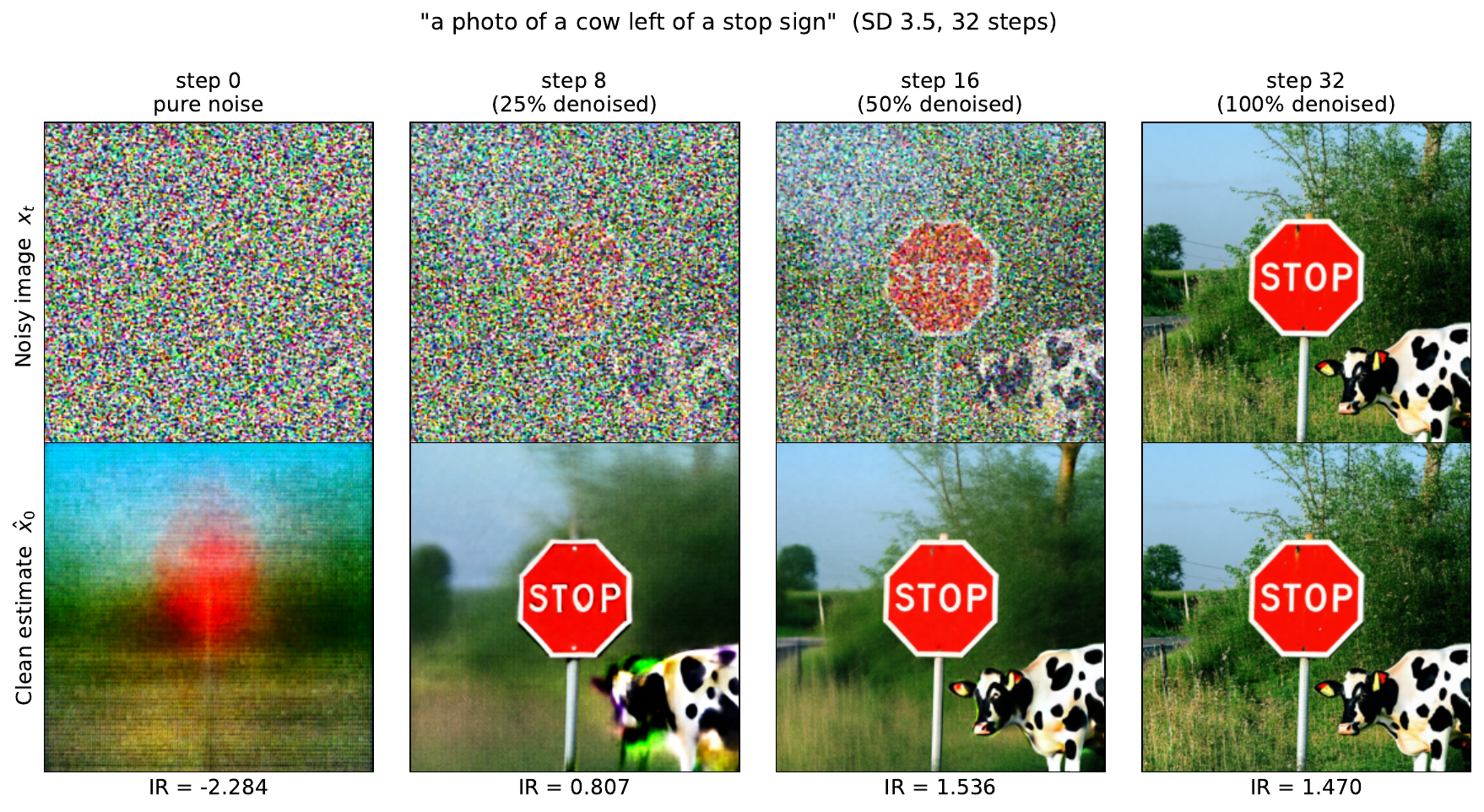}
    \caption{Predicted clean images $\hat{x}_0$ for Stable Diffusion 3.5 (32 steps).}
    \label{fig:sup_x0_sd35}
\end{figure*}

\clearpage

\newpage

\subsection{Generation Examples}
\label{sec:sup_generation}

Examples of neutral prompts from GenEval (prompt ids 0, 100, 200, 300, 400) under different models and sampling algorithms.

\begin{figure*}[h]
    \centering
    \includegraphics[width=\textwidth]{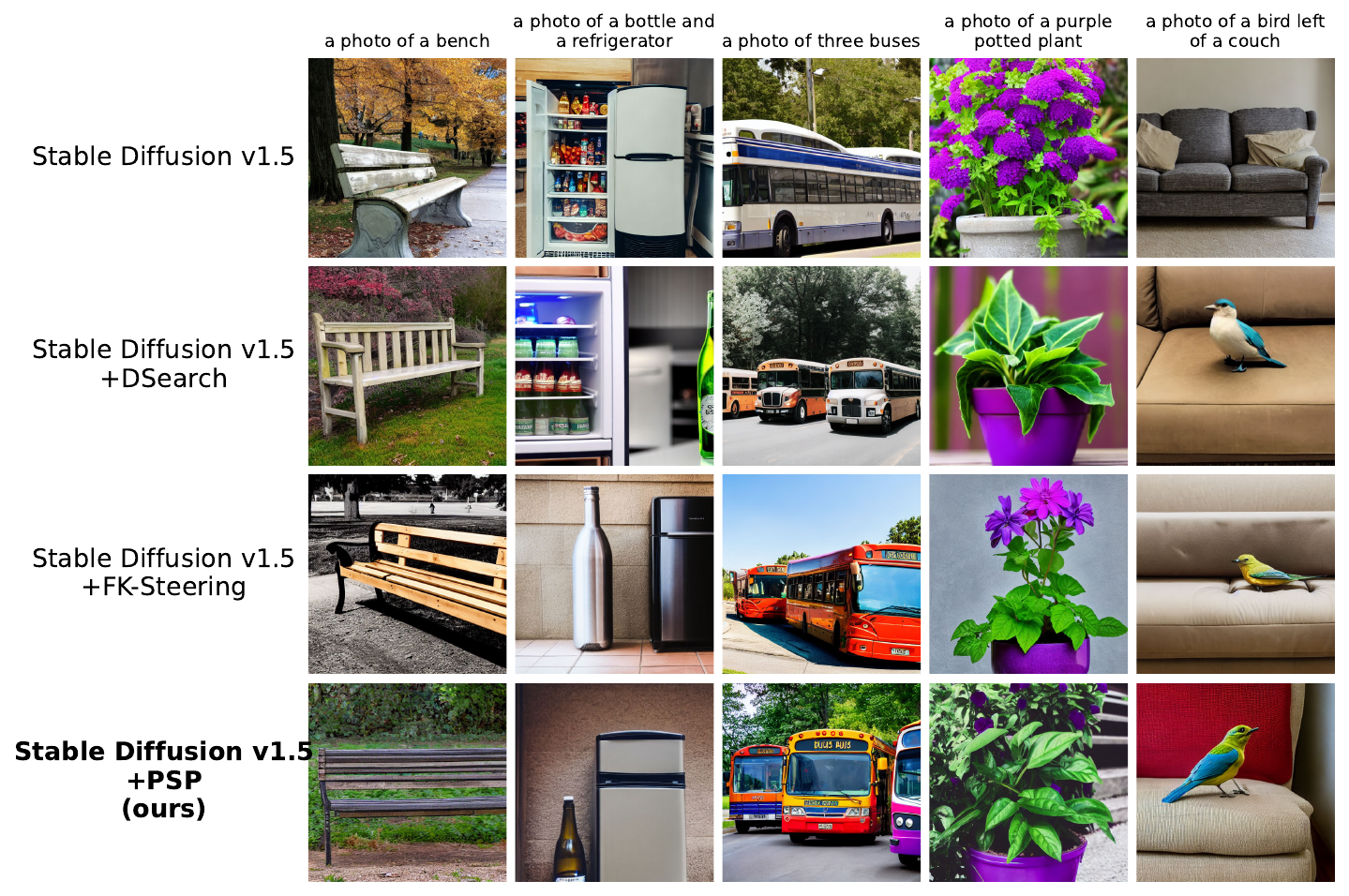}
    \caption{Generation examples for Stable Diffusion v1.5.}
    \label{fig:sup_generation_sd15}
\end{figure*}

\begin{figure*}[h]
    \centering
    \includegraphics[width=\textwidth]{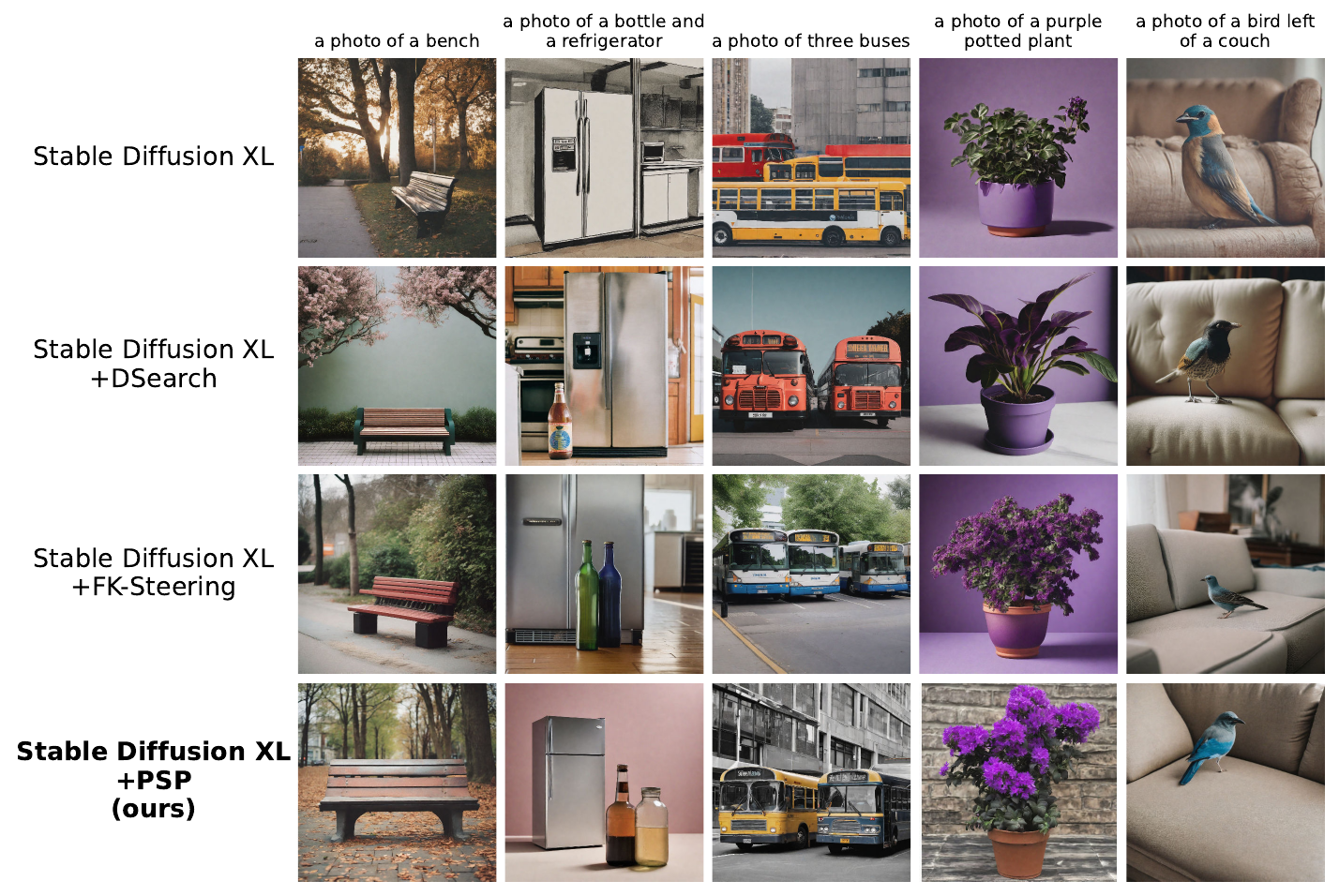}
    \caption{Generation examples for Stable Diffusion XL.}
    \label{fig:sup_generation_sdxl}
\end{figure*}

\begin{figure*}[h]
    \centering
    \includegraphics[width=\textwidth]{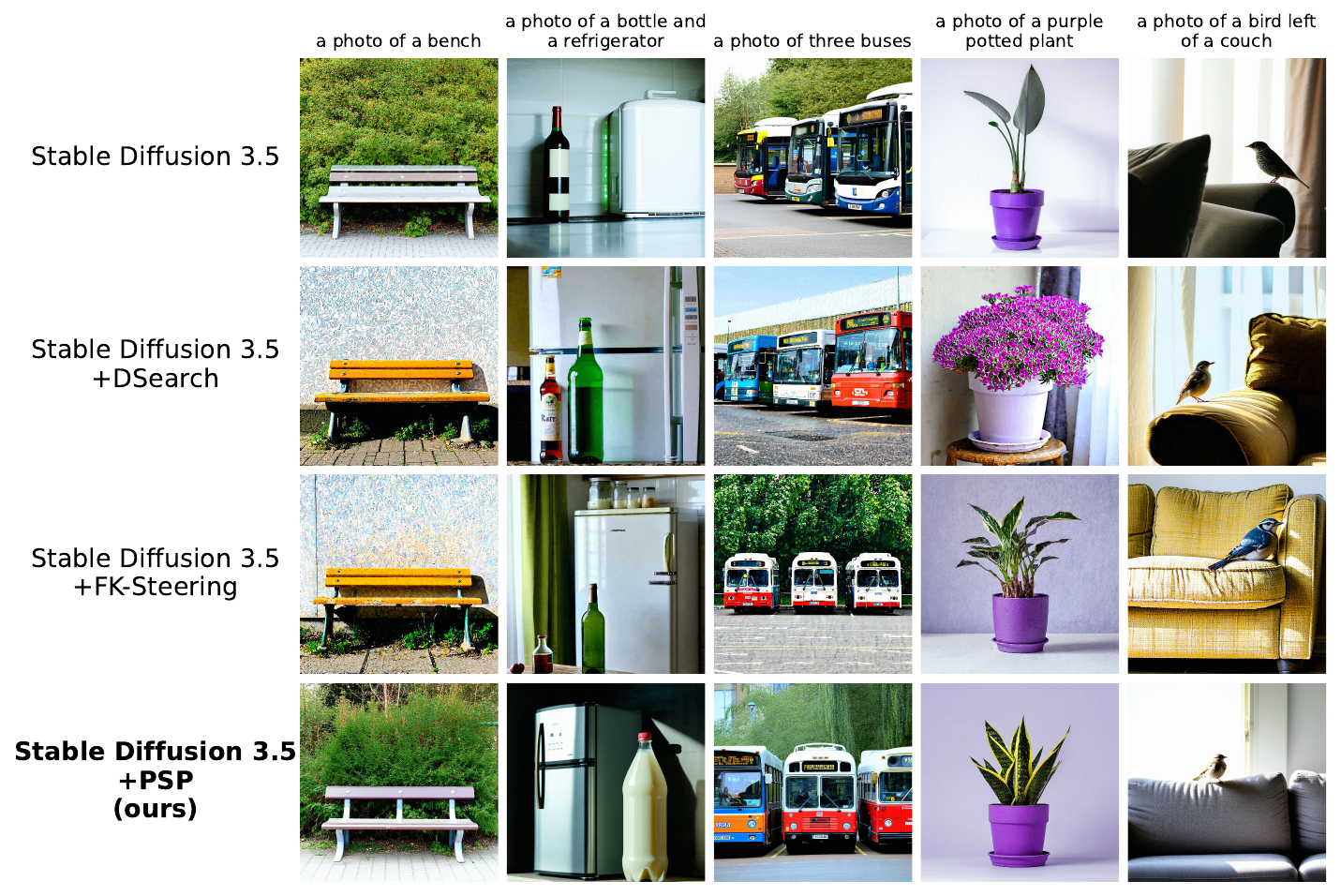}
    \caption{Generation examples for Stable Diffusion 3.5.}
    \label{fig:sup_generation_sd35}
\end{figure*}

\clearpage
\newpage

\subsection{Human Evaluations}
\label{sec:sup_human}

This section provides additional details on the human evaluation summarized in
\cref{sec:human_eval}, including recruitment, the annotation interface,
quality-control mechanisms, the assignment procedure, and score aggregation.
We adopted the same annotator instructions and evaluation criteria as
\cite{kamath_geneval_2025} so that our results are directly comparable to theirs
on the shared standard-sampler conditions. The instructions and evaluation
questions are therefore intentionally the same; however, the details of the
online platform and data-collection pipeline (interface, assignment logic, and
quality-control mechanisms described below) are our own, since these were not
publicly released for that work.

\paragraph{Recruitment and participants.}
Annotators were recruited through Prolific and were pre-screened to include only
workers with prior experience in AI-evaluation tasks. Participation was fully
anonymous: we stored only the Prolific-provided identifiers required to manage
submissions, and no personally identifying information was collected.
Participants were free to return the study at any time, and were instructed to
do so if they could not complete it in a single sitting. In total, 249
annotators contributed evaluations. Each annotator was assigned a session of
roughly 100 images and, based on our piloting, a session took approximately
25~minutes including the instructions.

\paragraph{Study organization.}
We ran the evaluation as a set of independent studies organized by backbone
(SD~1.5, SDXL, and SD~3.5). Studies were fully isolated
from one another: label availability, coverage requirements, and per-annotator
de-duplication were all enforced on a per-study basis, so no annotation in one
study could affect the counts or assignments of another. Every image (i.e.,
every prompt/method/backbone triplet, using a single seed) was configured to
receive exactly 3 independent evaluations.

\begin{figure}[h]
  \centering
  \includegraphics[width=0.9\linewidth]{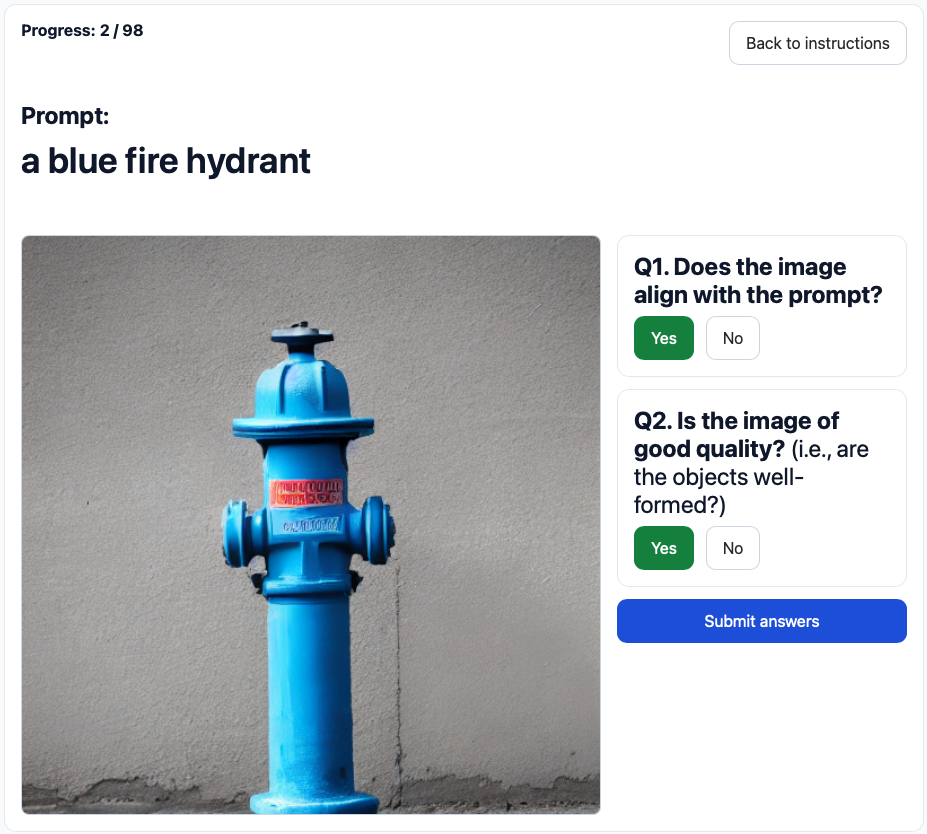}
  \caption{Example of an image presented for labeling, with the prompt shown
  above the image and the two binary (Yes/No) questions Q1 (alignment) and
  Q2 (quality). The answer buttons appear only after a short mandatory review
  delay.}
  \label{fig:human_exp_example}
\end{figure}

\paragraph{Annotation task and interface.}
The interface followed a fixed flow: (i) a consent screen, (ii) a multi-page set
of instructions, and (iii) the labeling task. For each image,
annotators answered two mandatory binary (Yes/No) questions:
\begin{itemize}
  \item \textbf{Q1 (alignment):} ``Does the image align with the prompt?''
  \item \textbf{Q2 (quality):} ``Is the image of good quality? (i.e., are the
        objects well-formed?)''
\end{itemize}
The prompt was shown above the image, and an example labeling screen is shown in
\cref{fig:human_exp_example}. The results reported in
\cref{tab:main_models_samplers} use the alignment question (Q1), which mirrors
the criterion used by the automated GenEval metric. We include the quality
question (Q2) to decorrelate perceived image quality from alignment: by giving
annotators a dedicated place to register quality concerns, we discourage them
from letting image quality influence their alignment judgment. Following
\cite{kamath_geneval_2025}, the quality responses themselves are not used in any
reported score and serve only this decorrelating purpose.

\paragraph{Quality control.}
To discourage rushed, low-effort responses, the Yes/No buttons for each image
were hidden and only revealed after a short mandatory review delay once the image
finished loading, and annotators were prompted to review the prompt and image
carefully during this period. Similarly, advancing through the instruction pages
was briefly locked on each page to encourage annotators to read them. Sessions
were subject to a fixed time budget (a non-rolling expiration measured from the
session start); annotators who exceeded it were timed out, and their in-progress
work was not counted toward the target. Per-image response times and
instruction/task timing were logged as telemetry for auditing.

\paragraph{Assignment and de-duplication.}
Assignments were computed server-side using a pre-materialized, slot-based
scheme. For each study we precomputed a fixed set of ``slots,'' where each slot
is a bundle of images sized to a single session ($\sim$100 images), and the
slots jointly cover every prompt/method/backbone triplet exactly 3 times. When
an annotator began, they atomically claimed a single unclaimed slot, guaranteeing
that (a) no annotator ever labels the same triplet more than once and (b) the
total number of evaluations per image is capped at the target. If a session
expired or was returned, its slot was released and recycled to a new annotator,
and only responses from completed slots were counted, which prevents
over-counting from partially finished sessions.

\paragraph{Aggregation and agreement.}
For each image we aggregated the 3 alignment (Q1) judgments by majority vote to
obtain a binary aligned/not-aligned decision, and method/backbone scores are the
mean of these per-image decisions. In total, the 8,295 images
(553 prompts $\times$ 5 methods $\times$ 3 backbones) received 24,885 ratings. As
a measure of reliability, we report the inter-annotator agreement rate, defined
as the fraction of images that received at least one vote for which all raters
gave the same answer; across the study this agreement was 80.3\%.

\paragraph{Validation against prior work.}
As noted in \cref{sec:human_eval}, restricting to the standard sampler on the
backbones shared with \cite{kamath_geneval_2025} reproduces their reported human
scores closely (SDXL: 0.531 vs.\ 0.566; SD~3.5: 0.787 vs.\ 0.770), which
validates that our recruitment, interface, and aggregation faithfully follow
their protocol.

\newpage

\begingroup
\centering
\captionof{figure}{Full set of instructions presented to annotators on the online
platform (in order), covering the task description, the two
evaluation questions, and worked examples of aligned vs.\ misaligned and high-
vs.\ low-quality generations. These instructions follow those of
\cite{kamath_geneval_2025}.}
\vspace{0.5in}
\includegraphics[width=0.7\textwidth]{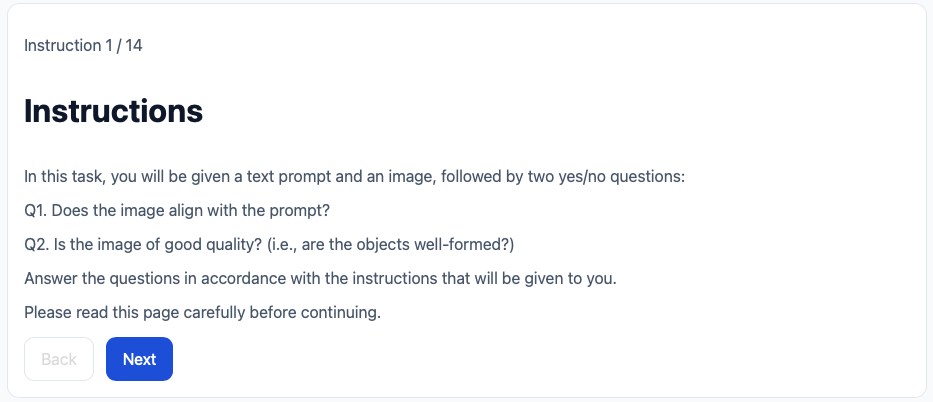}\par\medskip
\includegraphics[width=0.7\textwidth]{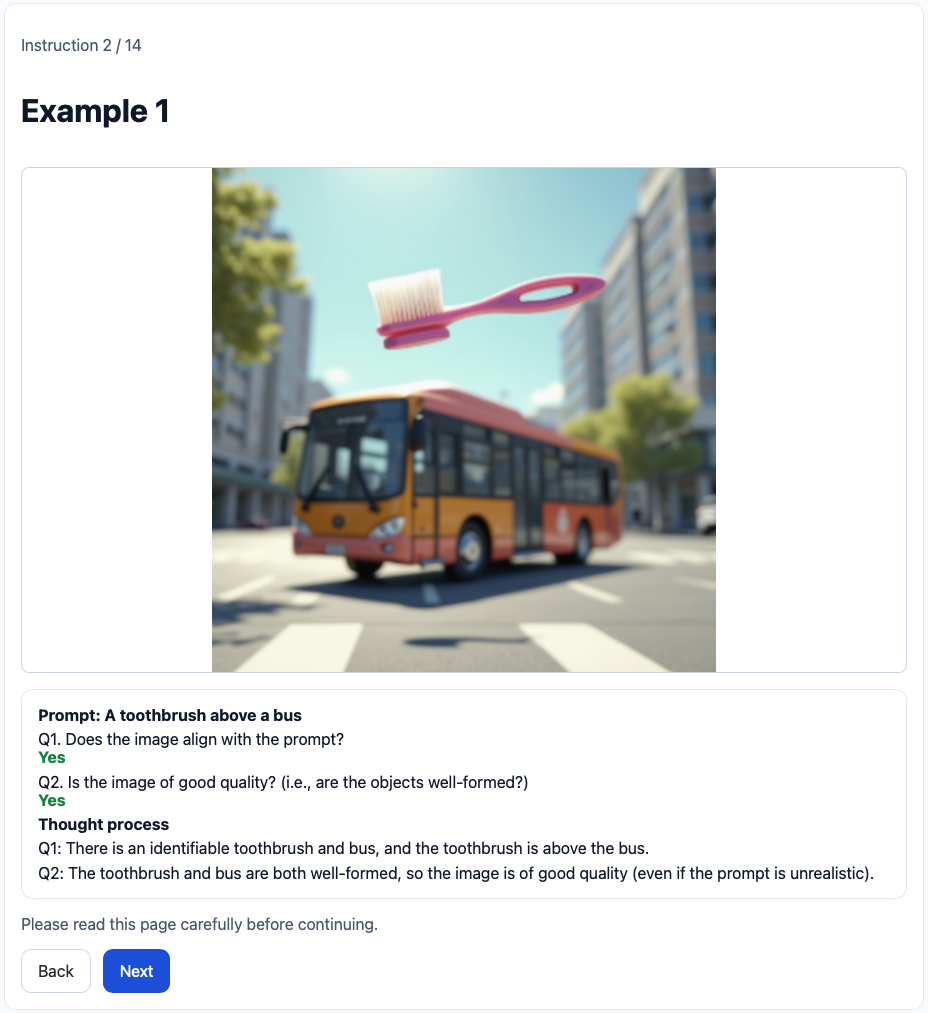}\par\medskip
\includegraphics[width=0.7\textwidth]{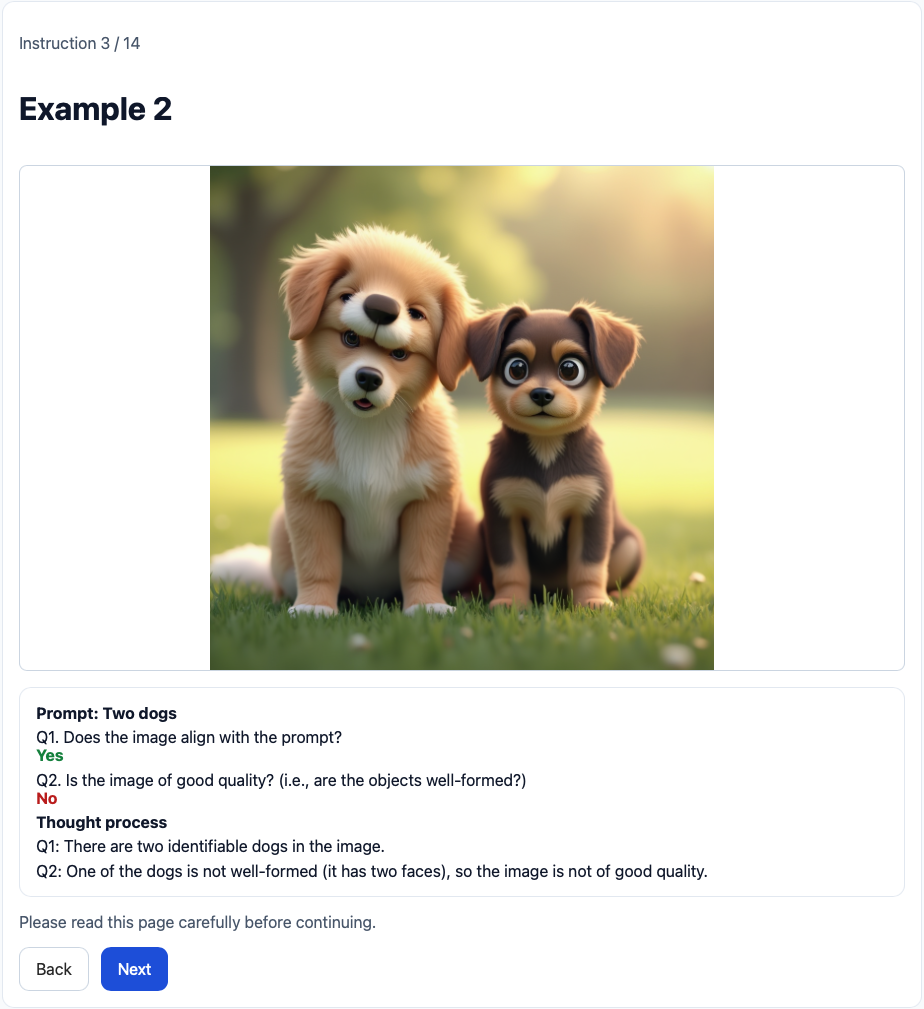}\par\medskip
\includegraphics[width=0.7\textwidth]{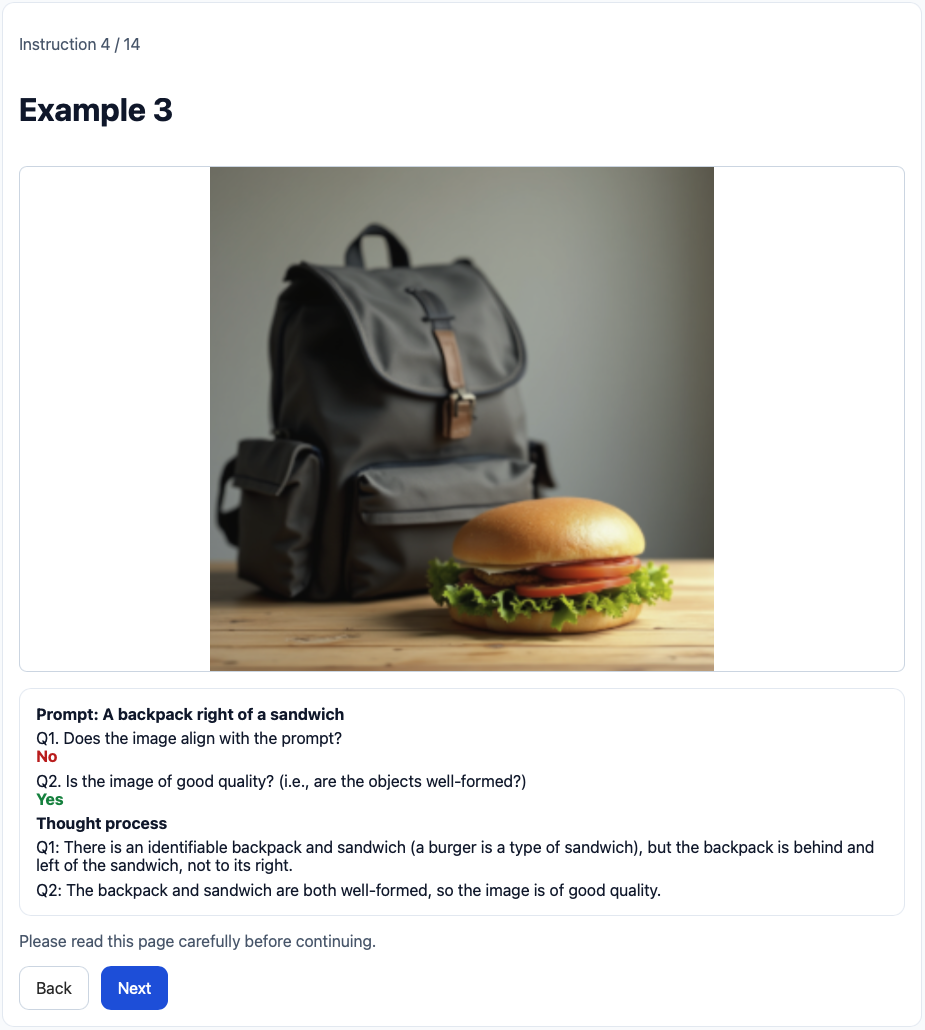}\par\medskip
\includegraphics[width=0.7\textwidth]{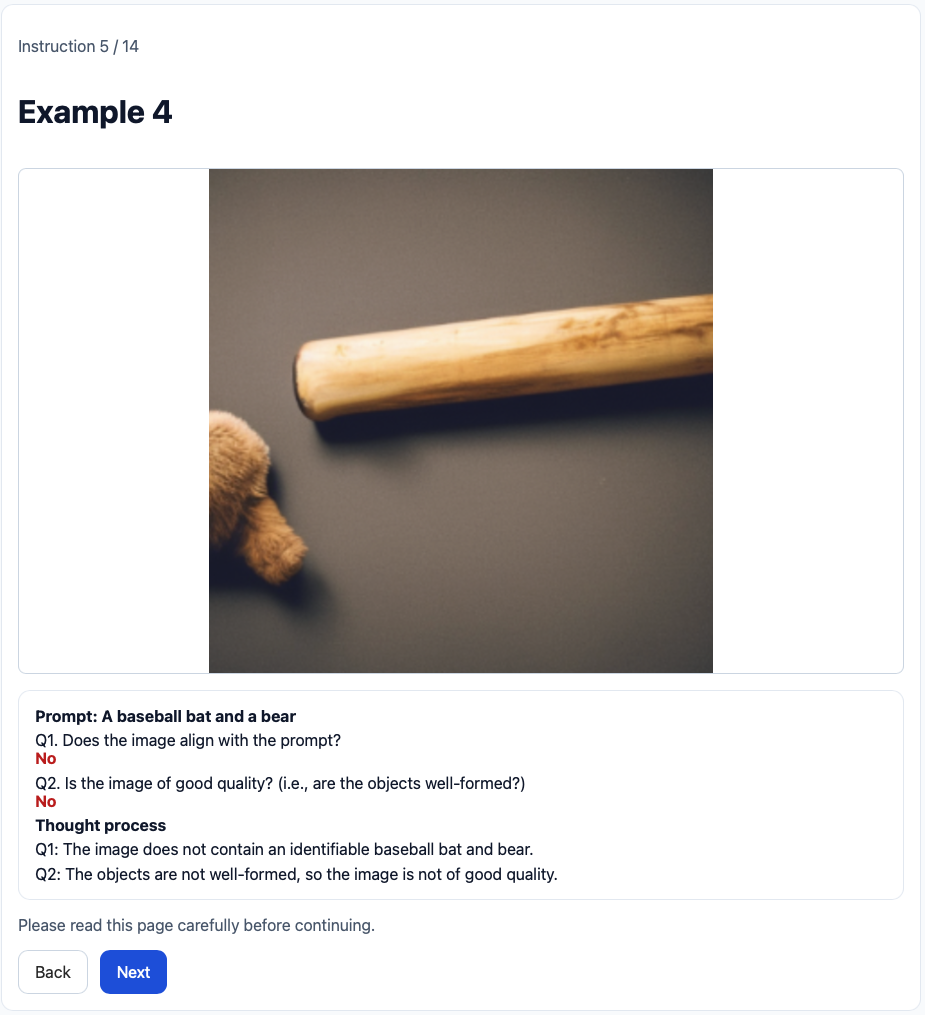}\par\medskip
\includegraphics[width=0.7\textwidth]{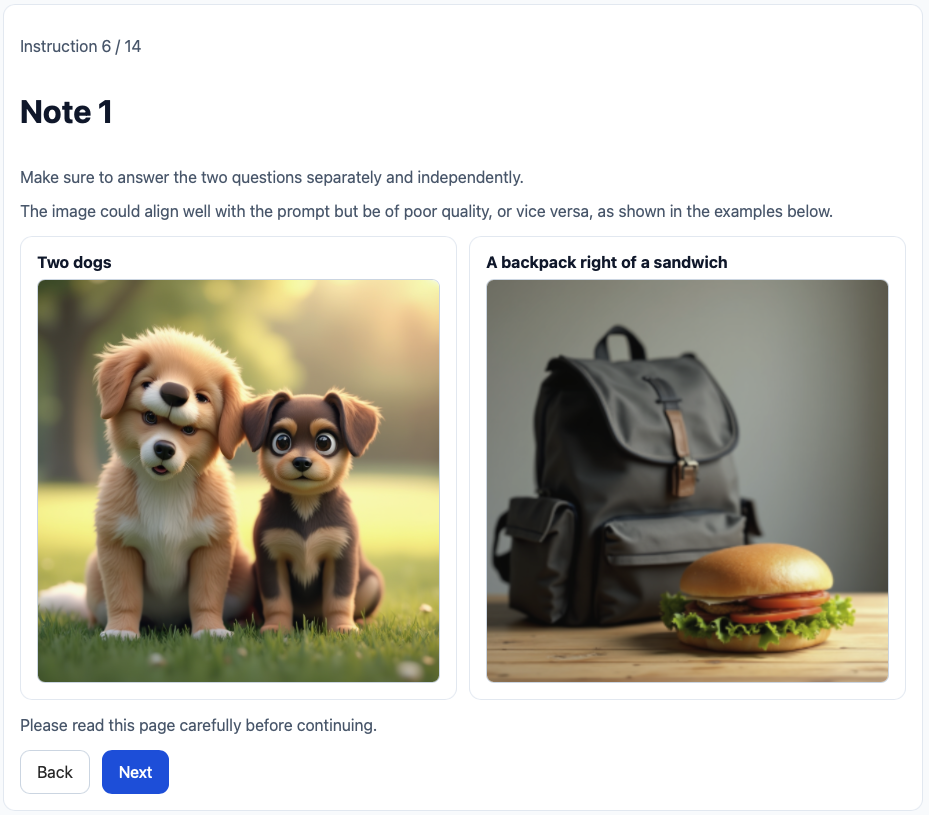}\par\medskip
\includegraphics[width=0.7\textwidth]{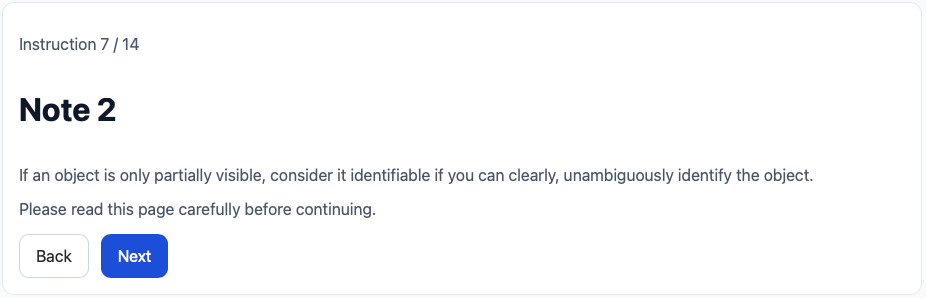}\par\medskip
\includegraphics[width=0.7\textwidth]{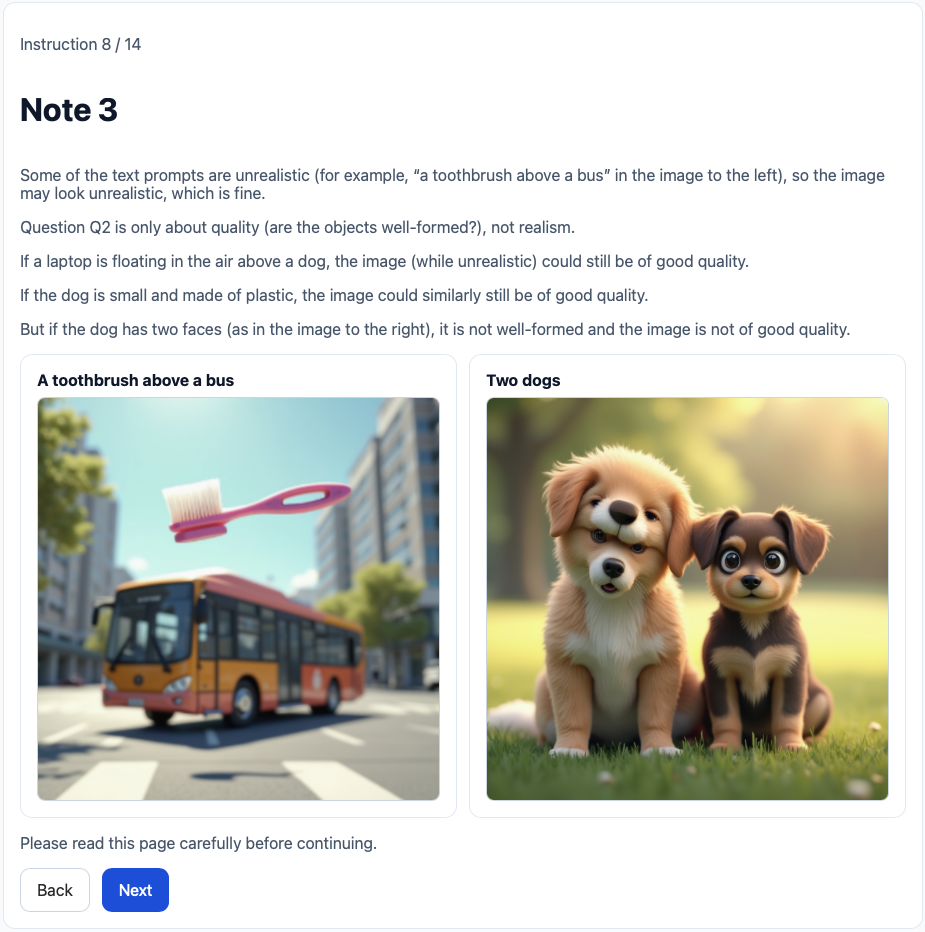}\par\medskip
\includegraphics[width=0.7\textwidth]{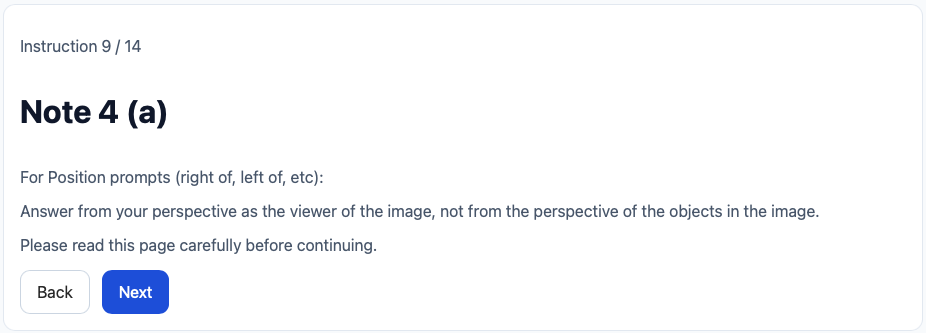}\par\medskip
\includegraphics[width=0.7\textwidth]{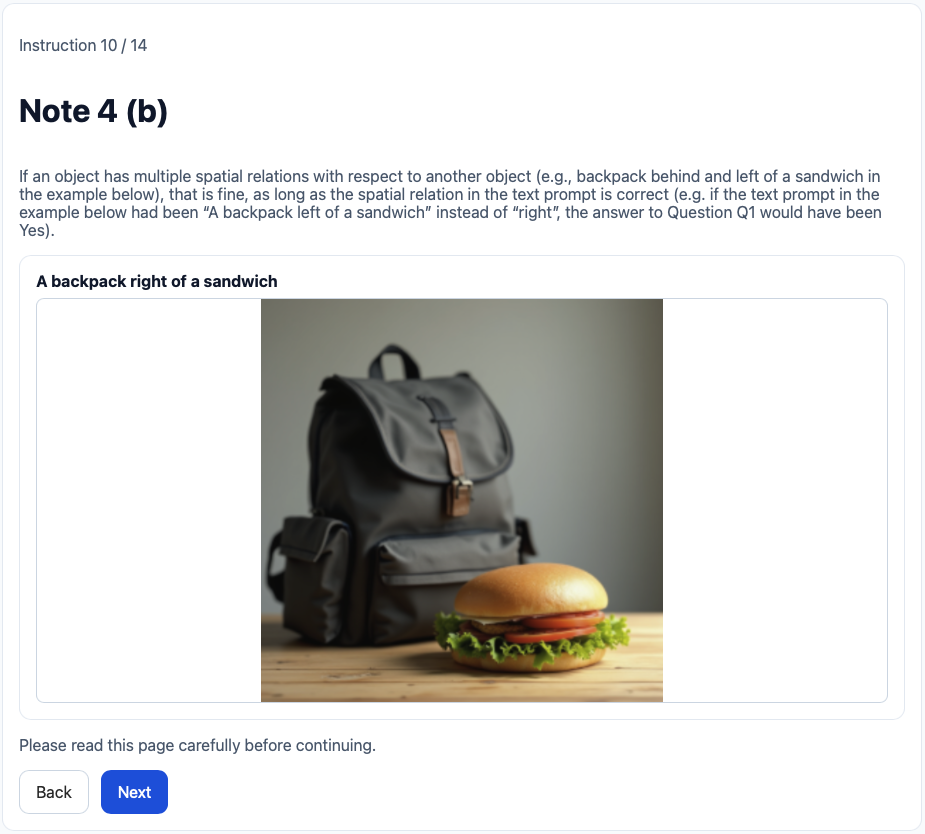}\par\medskip
\includegraphics[width=0.7\textwidth]{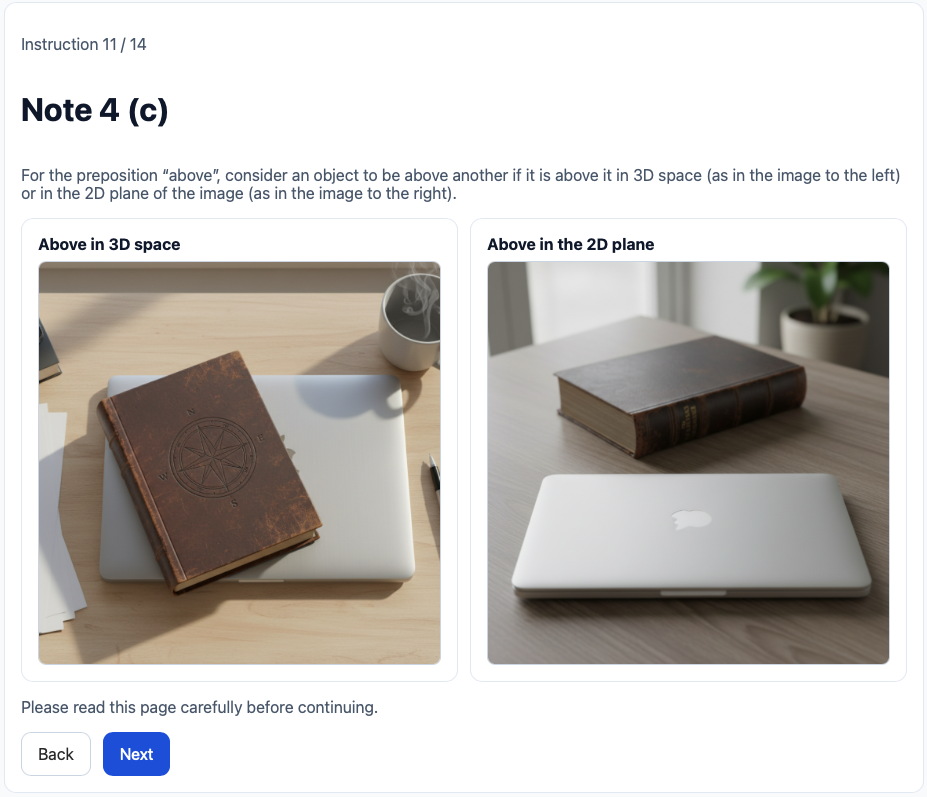}\par\medskip
\includegraphics[width=0.7\textwidth]{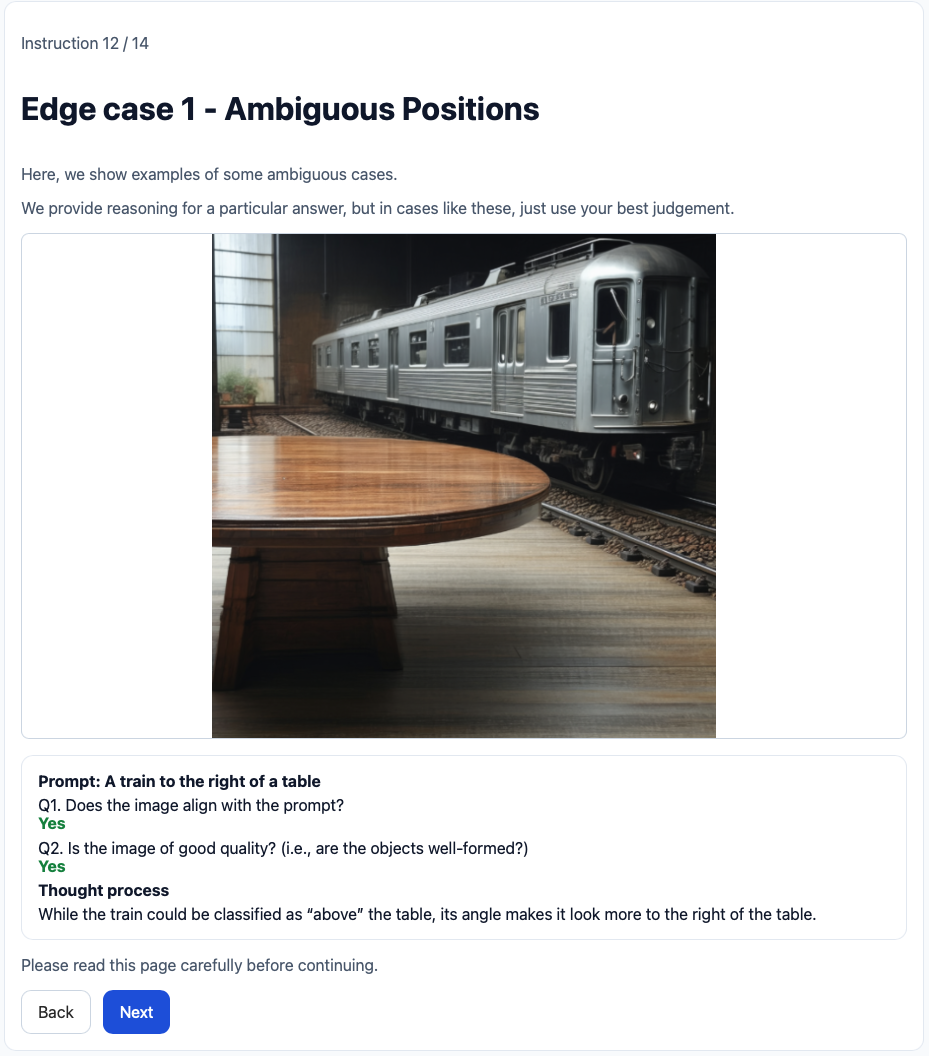}\par\medskip
\includegraphics[width=0.7\textwidth]{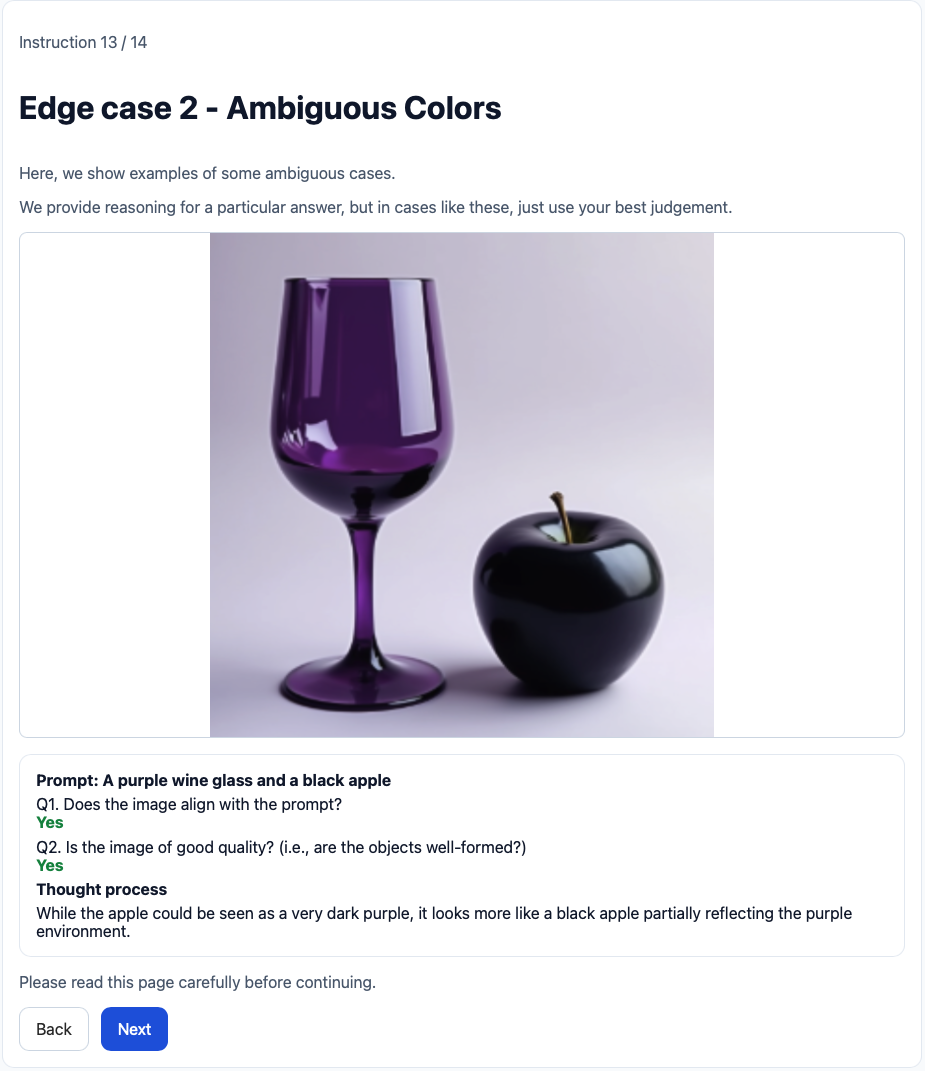}\par\medskip
\includegraphics[width=0.7\textwidth]{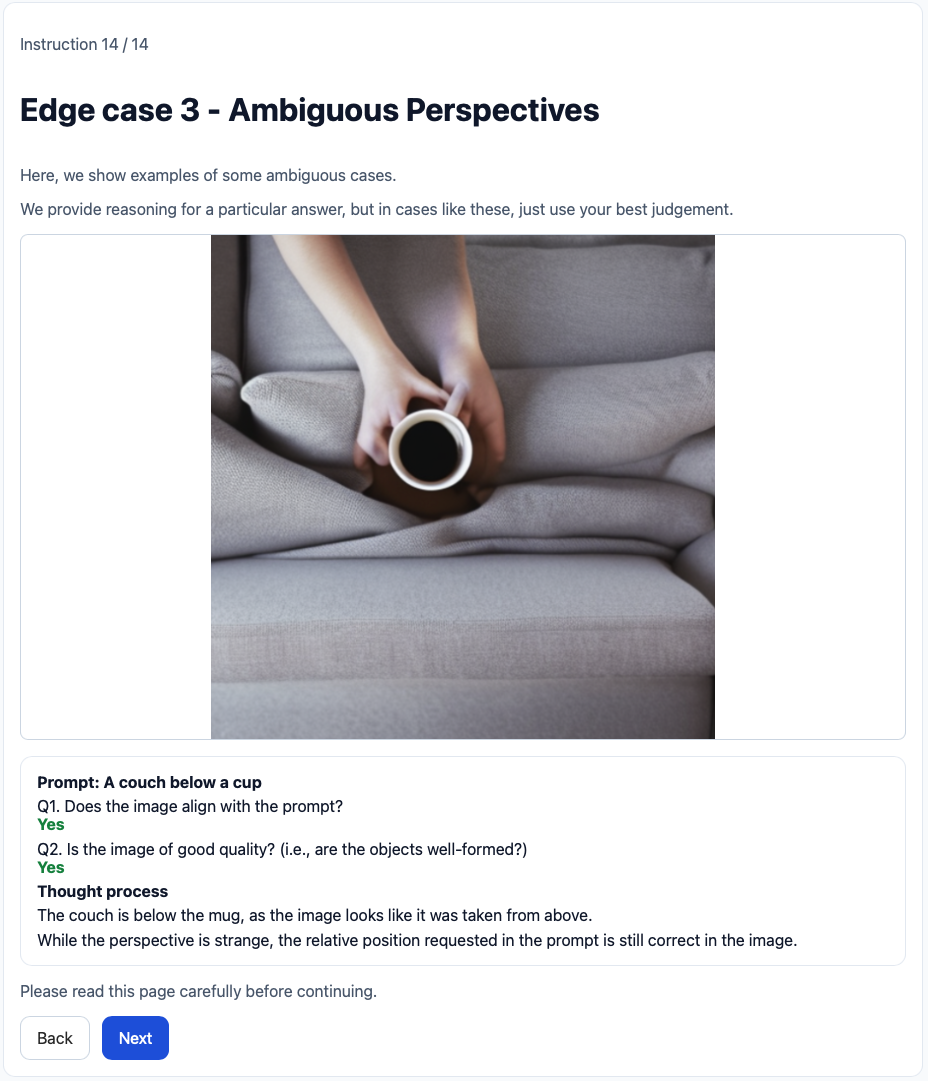}\par\medskip
\includegraphics[width=0.7\textwidth]{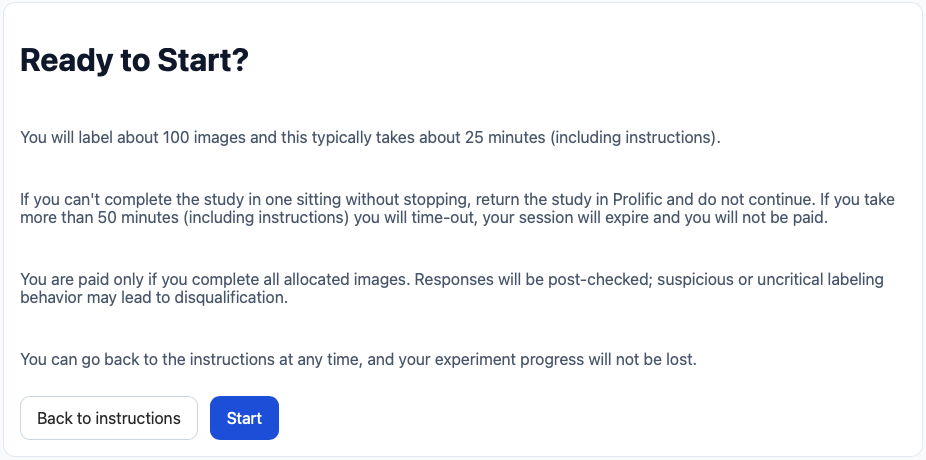}\par\medskip
\label{fig:human_exp_instructions}
\endgroup

\clearpage
\newpage

\end{document}